\pdfoutput=1

\documentclass[11pt]{article}

\usepackage[]{acl}

\usepackage{times}
\usepackage{latexsym}
\usepackage{booktabs}
\usepackage{multirow}
\usepackage{amssymb}
\usepackage{times}
\usepackage{amsmath}
\usepackage{latexsym}
\usepackage{graphicx}
\usepackage{algorithmic}
\usepackage{algorithm}
\usepackage{subfigure}
\usepackage{xcolor}
\usepackage{pifont}
\usepackage{textcomp}
\usepackage[T1]{fontenc}

\usepackage[utf8]{inputenc}

\usepackage{microtype}

\usepackage{inconsolata}

\usepackage{graphicx}

\usepackage{tabularx}

\usepackage{CJK}

\title{Scaling Laws for Fact Memorization of Large Language Models}

\author{
Xingyu Lu\textsuperscript{1}\thanks{\ \ Equal Contribution}\ , Xiaonan Li\textsuperscript{1}$^*$, Qinyuan Cheng\textsuperscript{1}, Kai Ding\textsuperscript{2}, Xuanjing Huang\textsuperscript{1}, Xipeng Qiu\textsuperscript{1}\thanks{\ \ Corresponding Author} \\
\textsuperscript{1}Fudan University,
\textsuperscript{2}INTSIG\\
luxy23@m.fudan.edu.cn, \ \{lixn20, xpqiu\}@fudan.edu.cn
}

\begin{document}
\maketitle
\begin{abstract}
Fact knowledge memorization is crucial for Large Language Models (LLM) to generate factual and reliable responses. However, the behaviors of LLM fact memorization remain under-explored.
In this paper, we analyze the scaling laws for LLM's fact knowledge and LLMs' behaviors of memorizing different types of facts.
We find that LLMs' fact knowledge capacity has a linear and negative exponential law relationship with model size and training epochs, respectively. Estimated by the built scaling law, memorizing the whole Wikidata's facts requires training an LLM with 1000B non-embed parameters for 100 epochs,  suggesting 
that using LLMs to memorize all public facts is almost implausible for a general pre-training setting.
Meanwhile, we find that LLMs can generalize on unseen fact knowledge and its scaling law 
is similar to general pre-training.
Additionally, we analyze the compatibility and preference of LLMs' fact memorization. For compatibility, we find LLMs struggle with memorizing redundant facts in a unified way. Only when correlated facts have the same direction and structure, the LLM can compatibly memorize them. This shows the inefficiency of LLM memorization for redundant facts.
For preference, the LLM pays more attention to memorizing more frequent and difficult facts, and the subsequent facts can overwrite prior facts' memorization,
which significantly hinders low-frequency facts memorization. 
Our findings reveal the capacity and characteristics of LLMs' fact knowledge learning, which provide directions for LLMs' fact knowledge augmentation.
\end{abstract}
\vspace{-10pt}
\section{Introduction}
\vspace{-5pt}

Large Language Models (LLM) have demonstrated remarkable abilities over a wide range of tasks~\citep{gpt4,llama2,gemini_1.5,qwen,deepseek_v2,internlm2,Sun2024MOSS}. However, LLMs are prone to generating non-factual and fabricated contents, which is usually called ``hallucination''~\citep{hallucination_survey_tencent, hallucination_survey_hit,hallucination_survey_usc} and undermines LLMs' reliability.

LLMs' factual responses highly rely on fact memorization.
Specifically, the LLM memorizes fact knowledge during pre-training and the subsequent fine-tuning enables it to extract corresponding fact knowledge for the given instruction~\citep{llm_physics_31}.
If the base LLM does not memorize specific knowledge, 
it will be challenging for the fine-tuned LLM to correctly answer the corresponding question~\citep{learning_or_self_aligning}.
Additionally, fine-tuning with unmemorized fact knowledge even encourages LLMs' hallucination~\citep{flame_fact_aware_alignment, finetuning_on_new_knowledge_encourages_hallucination}.
Despite the critical role of fact memorization, the behaviors of LLM fact memorization remain largely under-explored.
\begin{figure}[t]
    \centering
    \includegraphics[width=0.21\textwidth]{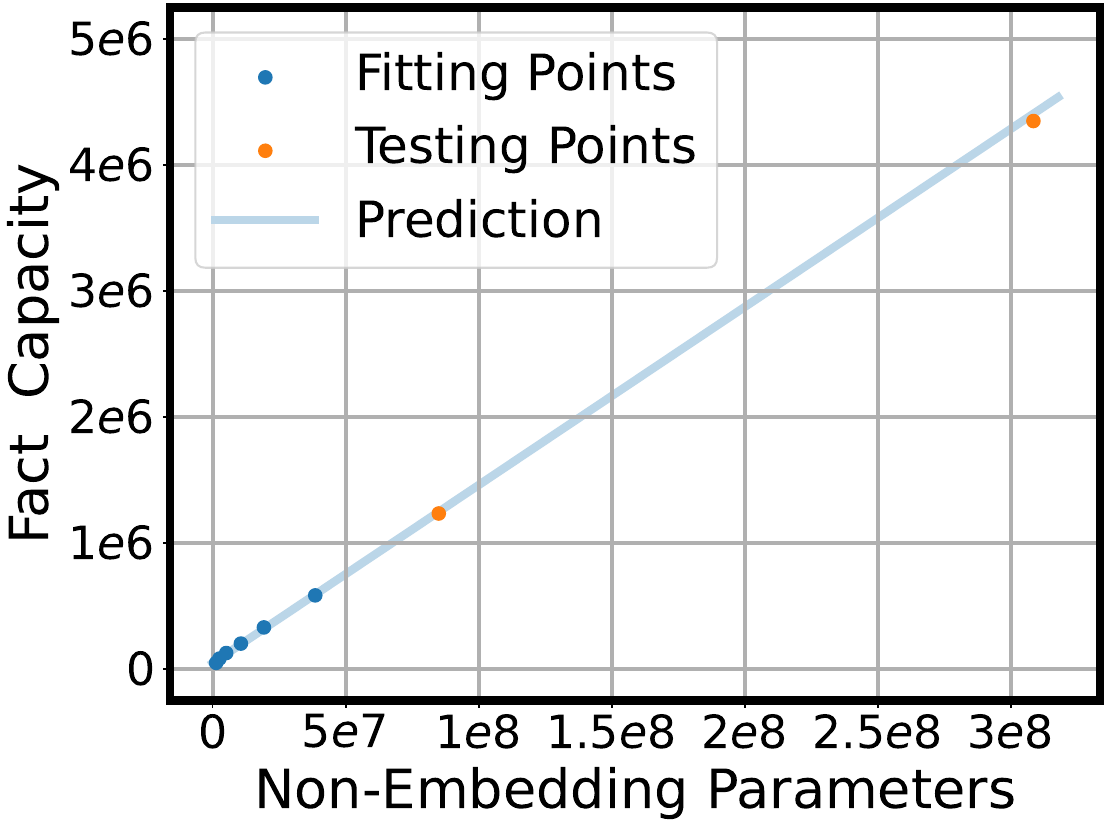}
    \includegraphics[width=0.215\textwidth]{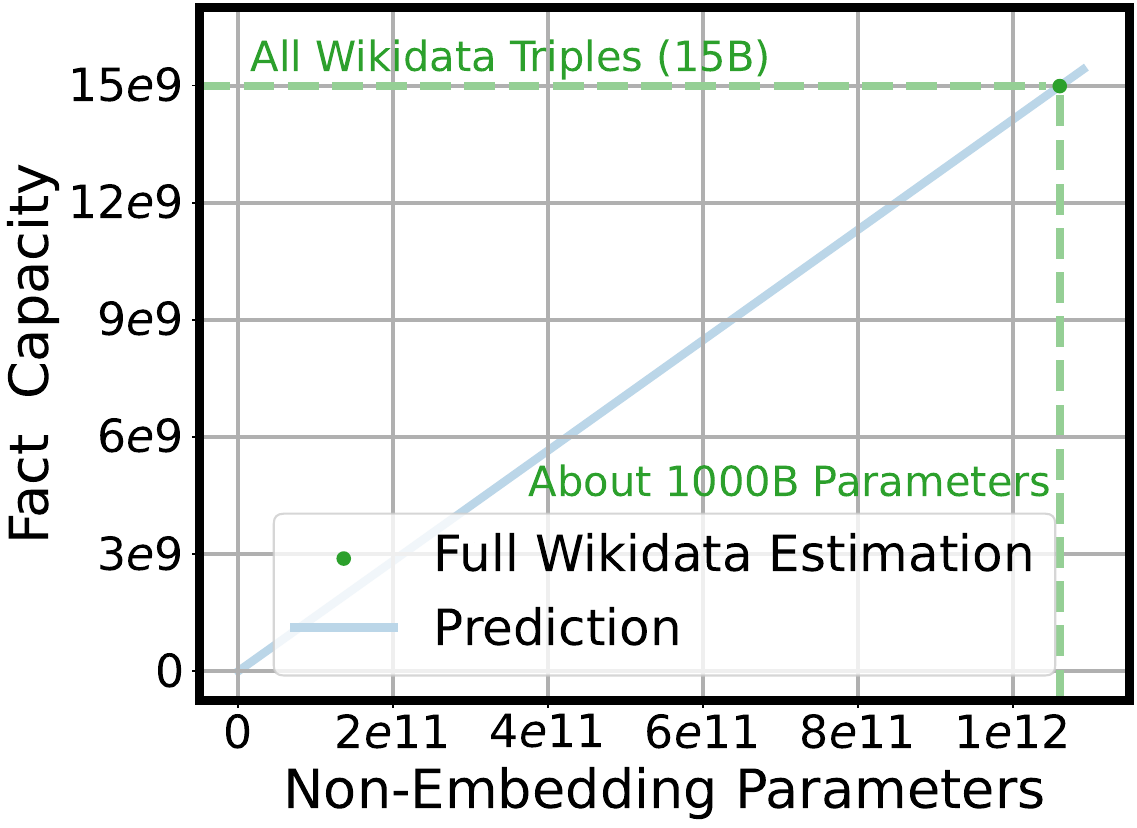}
    \vspace{-7pt}
    \caption{The fact capacity of LLMs with different sizes on Wikidata, under 100 training epochs. According to the predicted scaling law, memorizing all Wikidata triples (15B) requires 1000B non-embed parameters.}
    \label{fig:fact_capacity_and_parameter_Wikidata}
    \vspace{-18pt}
\end{figure}
Previous work usually analyzes the pre-trained LLMs' various abilities through the loss on unstructured text~\citep{scaling_law_for_lm, chinchilla_scaling_law}, and it is hard to reflect LLMs' fact memorization for two reasons:
1. The composition of pre-training corpus is highly complicated and fact knowledge appears in it mixedly and unevenly, which makes it hard to accurately quantify the fact knowledge in massive pre-training data.
2. The widely used metric, loss, can not directly measure the LLM fact memorization since not all tokens are fact-related.

This paper makes progress in quantitatively analyzing LLM fact memorization behaviors, including the scaling laws and behaviors of memorizing different types of facts. 
We focus on the memorization of atomic facts to facilitate accurately quantifying the number of facts and the memorization accuracy.
We define atomic fact knowledge as a (key, attribute, value) triple, e.g., (SpaceX, CEO, Elon Musk), following~\citet{llm_physics_33}. Given a key and an attribute, if the LLM correctly predicts the corresponding value, we consider it to memorize this fact knowledge.
In this way, we can accurately quantify the number of fact knowledge and whether the LLM fully memorizes a specific fact, which facilitates a more accurate quantitative analysis of LLM's fact memorization behaviors.

Based on this setting, we analyze the LLM's fact memorization behaviors on massive facts from a large real-world information table. Specifically, we analyze the fact memorization scaling law of LLMs and LLMs' behaviors of memorizing different types of fact knowledge, including the following research questions (RQ):
\vspace{-3pt}
\paragraph{RQ1: \textit{How does LLM's fact knowledge capacity scale with its size and training epochs?}}
We define the fact knowledge capacity as the maximum fact triple quantity that the LLM can accurately memorize. We find that LLM's fact capacity linearly scales with its size under the same training epochs. Additionally, we find that the training epochs required for LLMs to memorize fact knowledge is significantly larger than one and this leads to higher training cost than general knowledge learning in pre-training.
Increasing training epochs can initially increase the LLM's fact capacity and then reach saturation, which exhibits a trend of negative exponential law.
Additionally, we extend our experiments to the Wikidata and the results exhibit a consistent trend, shown in Figure~\ref{fig:fact_capacity_and_parameter_Wikidata}.
According to the scaling law, under 100 training epochs, memorizing all Wikidata's fact triples requires about 1000B non-embed parameters, which seems very costly. These indicate the necessity of supplementing LLMs with fact knowledge by external information, like Retrieval-Augmented Generation (RAG)~\citep{realm_rag, rag_survey, self_rag, hybrid_hierachical_rag, llatrieval, silo, replug}.
\vspace{-3pt}
\paragraph{RQ2: \textit{Can LLMs efficiently memorize redundant facts?}}
Many facts are derivable and thus redundant. For example, ``Ivanka is Trump's daughter'' can derive from ``Trump is Ivanka's father''.
We analyze whether LLMs can efficiently memorize redundant facts, i.e., whether LLMs can save memorization capacity when simultaneously memorizing redundant facts. 
We find that LLMs struggle with efficiently memorizing redundant information. 
In general cases, when memorizing the redundant and non-redundant information of the same scale, the LLM exhibits a similar memorization rate. Only under specialized conditions, e.g., the correlated facts have the same direction and structure, the LLM can efficiently memorize them.
These demonstrate LLMs' inefficiency in redundant fact memorization. 
Since massive redundant facts can appear in pre-training data in various forms, these indicate it is not cost-effective to use LLMs' parameters to store fact knowledge, and using a non-parametric method, like RAG, can be more efficient.

\vspace{-3pt}
\paragraph{RQ3: \textit{What influences LLM's memorization preference for different types of fact knowledge?}}
During pre-training, LLMs meet various facts and only memorize portions of them. We analyze LLMs' fact memorization preference in three aspects: frequency, difficulty and memorization order. We find that LLMs pay more attention to memorizing more frequent and difficult facts. 
Additionally, when an LLM memorizes two types of facts sequentially, the subsequent facts will significantly overwrite the memorization of prior facts.
These further explain LLMs' inferior memorization of low-frequency facts since they appear infrequently during pre-training process and thus can be easily overwritten by subsequent pre-training knowledge.

Beyond fact memorization, we also analyze an interesting topic of fact knowledge generalization:
\vspace{-3pt}
\paragraph{RQ4: \textit{Can LLMs generalize on unseen fact knowledge?  What is the relation between fact memorization and generalization?}}
Surprisingly, we find that the LLM can generalize on unseen facts to a certain level and its scaling law is highly similar to common pre-training LLM scaling law~\citep{scaling_law_for_lm}.
The generalization accuracy is determined by the type of fact and some types of facts exhibit high generalizability, suggesting the potential of improving LLMs' factuality by adaptively leveraging fact generalization. 
Meanwhile, we find a qualitative relation between fact memorization and generalization: To the same type of fact, the easier the LLM is to memorize it, the better the LLM generalizes on the unseen set. 
This indicates that both LLM fact memorization and generalization are based on the correlation between input and output~\citep{short_cut_learning}.
If there is a stronger correlation between the input and output of one type of fact, it will be easier for the LLM to memorize and learn about the type of fact knowledge in a unified manner. Conversely, if the correlation is minimal, LLM needs to memorize facts individually, and is hard to generalize on unseen ones.

We summarize our contribution as follows: 1) To the best of our knowledge, this paper is the first to quantitatively analyze LLMs' scaling laws and behaviors of fact memorization on massive real-world facts. 2) Our findings reveal the capacity and characteristics of LLMs' fact knowledge learning. 
These results show that LLMs are highly inefficient for fact memorization from multiple perspectives, which suggests leveraging non-parametric methods, e.g., RAG, to enhance the fact knowledge of LLMs. 3) We find that LLMs can generalize on unseen facts and different types of facts show different generalizability, which indicates the potential of improving LLMs' factuality by adaptively leveraging LLMs' fact generalization. 4) We release our code to facilitate future research\footnote{\href{https://github.com/StarLooo/Scaling\_Law\_LLM\_Fact\_Memorization}{https://github.com/StarLooo/Scaling\_Law\_LLM\_Fact\_\\Memorization}}.

\vspace{-4pt}
\section{Preliminary}
\vspace{-4pt}
In this paper, we focus on the quantitative analysis of LLMs' atomic fact knowledge memorization and we introduce the experiment setup as follows.

\paragraph{Atomic Fact Knowledge Memorization}
We define atomic fact knowledge as a (key, attribute, value) triple, e.g., (SpaceX, CEO, Elon Musk), and we cast fact memorization as a triple value prediction task.
Specifically, for a fact triple ($k$, $a$, $v$), we use the cross-entropy loss to train the LLM to predict the value by the ($k$, $a$) as:
\begin{align}
    p = \operatorname{LLM}(template_a(k,a)),
    \label{eq:value_prediction}
\end{align}
where $k$ and $a$ are the key's name and attribute name, and $template_a$ is the natural language template of the attribute to make the LLM's input more coherent for realism. We adopt one template for one attribute for simplicity. Our pilot experiments show that various numbers of templates lead to consistent results, shown in Appendix~\ref{app:template_quantity_effect}. Since we focus on fact memorization, we use the same input for training and inference. 

After training on facts $D=\{k_i,a_i,v_i\}_{i=1}^{|D|}$, we evaluate the LLM's \textbf{Memorization Rate (MR)} as:
\begin{align}
    \operatorname{MR}(D) = \operatorname{average}_{i=1}^{|D|}(\operatorname{EM}(p_i,v_i)),
\end{align}
where $\operatorname{EM}$ means exact match, and $p_i$ and $v_i$ are the $i$-th fact's prediction and value. In this way, we can use the memorization rate to accurately quantify the portion of facts the LLM has memorized.

\paragraph{Dataset}
\begin{table}[]
\small
\centering
\begin{tabular}{@{}lll@{}}
\toprule
\textbf{Field}             & \textbf{Description}    & \textbf{Example}   \\ \midrule
Company$^*$ & company name            & Tiktok Co., Ltd.   \\
Credit-No                  & social credit number      & 91110105MA...      \\
Operator                   & legal representative    & Lidong Zhang       \\
Start-Date                & founding date           & 2003.11.2      \\
Title                      & representative title & Executive Director \\
Type                       & company type            & Co., Ltd.               \\
Register-Capital & registered capital & \textyen 10$^5$ \\
Longitude                  & company longitude       & 116.497976         \\
...                        & ...                     & ...                \\ \bottomrule
\end{tabular}
\vspace{-5pt}
\caption{Company information table, which has 22 fields and 10M lines. ``Company$^*$'' is the primary key.
The information of overall fields is shown in Appendix~\ref{app:overall_table_information}.}
\label{tab:table_attribute_example}
\vspace{-4pt}
\end{table}
This paper mainly conducts experiments on massive facts of a large real-world company information table,
which is provided by a commercial data company, INTSIG\footnote{INTSIG is a leading company of intelligent document recognition. \href{https://www.intsig.com/}{https://www.intsig.com/}}.
The table contains various attributes of massive companies and we use facts like (Company, Attribute, Value) for experiments.
The involved facts are from the real world and the types of them are diverse, and thus closely mirror the various facts in pre-training process.
We show the table's statistics and sample row in Table~\ref{tab:table_attribute_example}. 
Additionally, experiments on Wikidata also show consistent trends (Section~\ref{sec:scaling_law}).
\vspace{-5pt}
\paragraph{Implementation Details}
We mainly use the model architecture and tokenizer of Qwen~\citep{llm_qwen}. We conduct experiments over various sizes of LLMs from 20M to 0.5B. 
We mainly train LLMs' fact memorization from scratch and we show the results on pre-trained LLMs in Appendix~\ref{app:pre_training_effect}.
For the specific hyper-parameters of each model size and overall implementation details, please refer to Appendix~\ref{app:implementation_details}.
\vspace{-4pt}
\section{Fact Capacity Scaling Laws}
\begin{figure}[]
    \centering
    \includegraphics[width=0.32\textwidth]{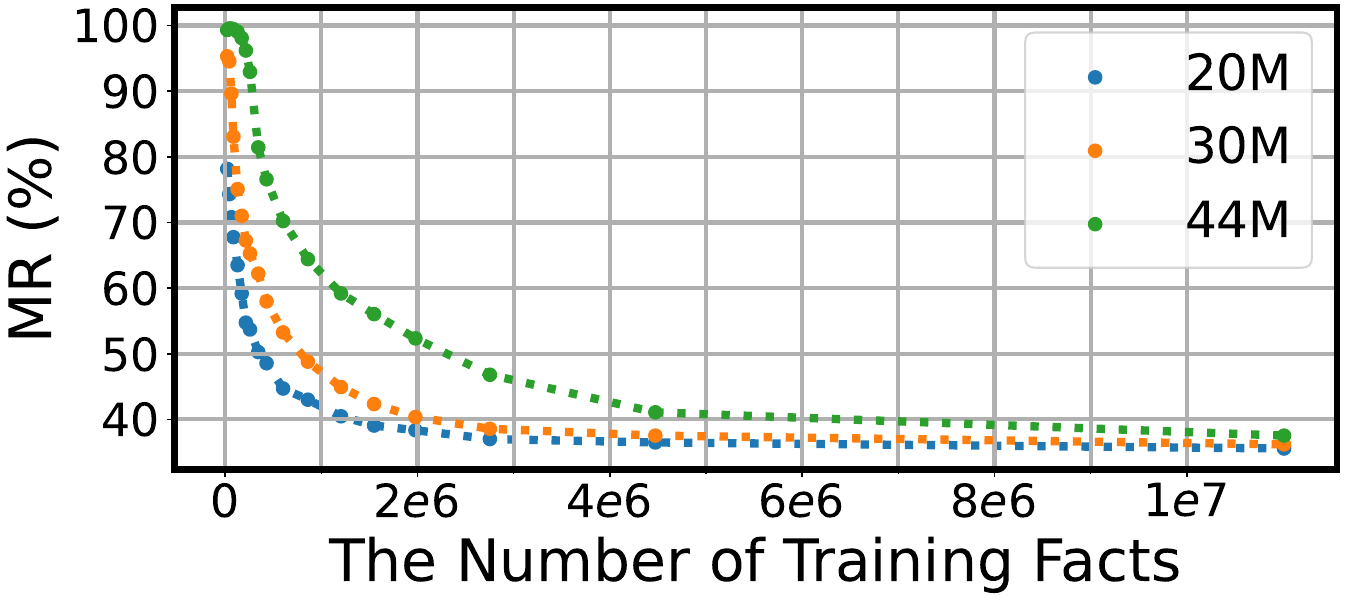}
    \vspace{-5pt}
    \caption{LLMs' memorization rate under different numbers of training facts.}
    \label{fig:mr_of_different_fact_quantity}
    \vspace{-10pt}
\end{figure}

\vspace{-4pt}
\label{sec:scaling_law}
\paragraph{Exploratory Experiment} 
First, we observe the same LLM's memorization rate over varying numbers of training facts under the same training epochs. We show the results in Figure~\ref{fig:mr_of_different_fact_quantity}. We see that the memorization rate significantly decreases with the increasing facts.  
These initially show that there is a memorization capacity upper limit for the LLM with the same size and training epochs.

In this section, we explore the scaling laws of LLMs' fact capacity. We define the fact capacity as the maximum fact quantity that the LLM can accurately memorize as:
\begin{align}
C = 
\max(|D|)  \text{ s.t. } \operatorname{MR}(D) > \phi\%,
\end{align}
where $D$ is training facts, a list of randomly sampled facts from all facts, and $\phi$ means a high MR close to 100\%. In experiments, we set $\phi\%$ to be 95\% and enumerate $D$s of varying sizes to find the maximum $|D|$ that MR($D$) is between $[\phi\%, (\phi + 1)\%]$.
\paragraph{Scaling Law of Fact Capacity and Model Size}
\begin{figure}[]
    \centering
    \subfigure[50 Epochs]{
    \includegraphics[width=0.23\textwidth]{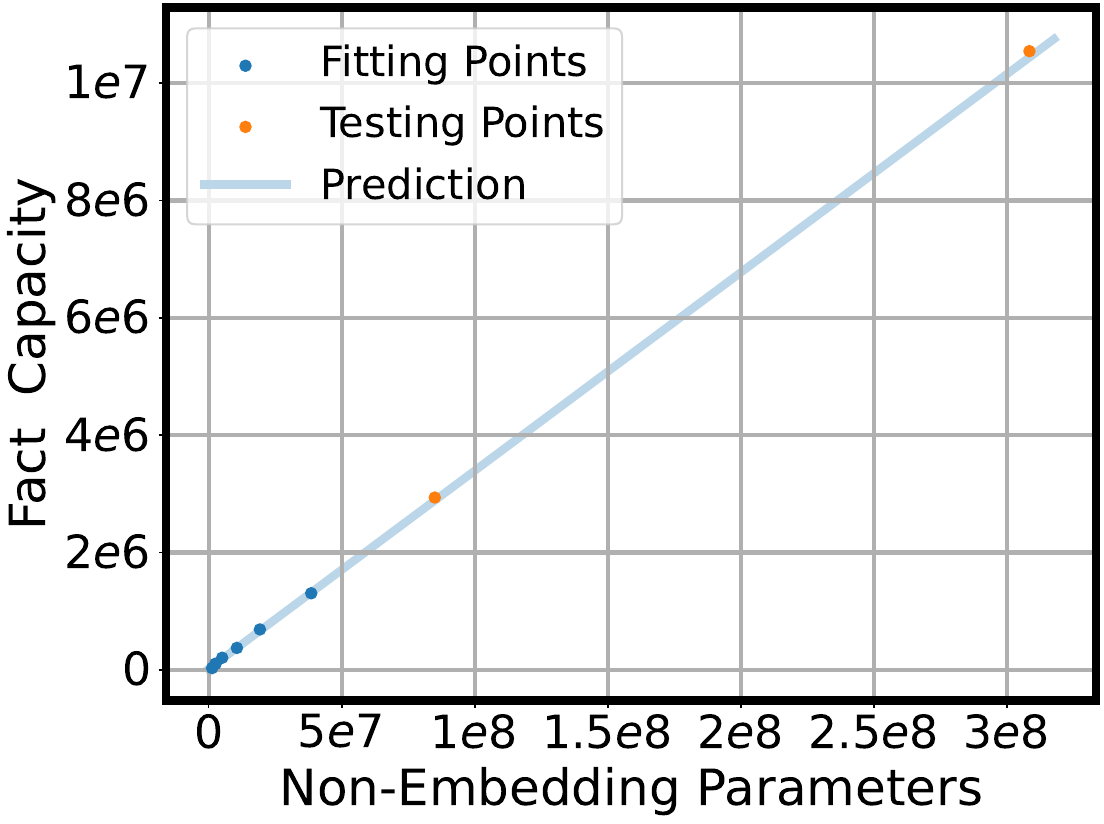}}
    \subfigure[200 Epochs]{
    \includegraphics[width=0.23\textwidth]{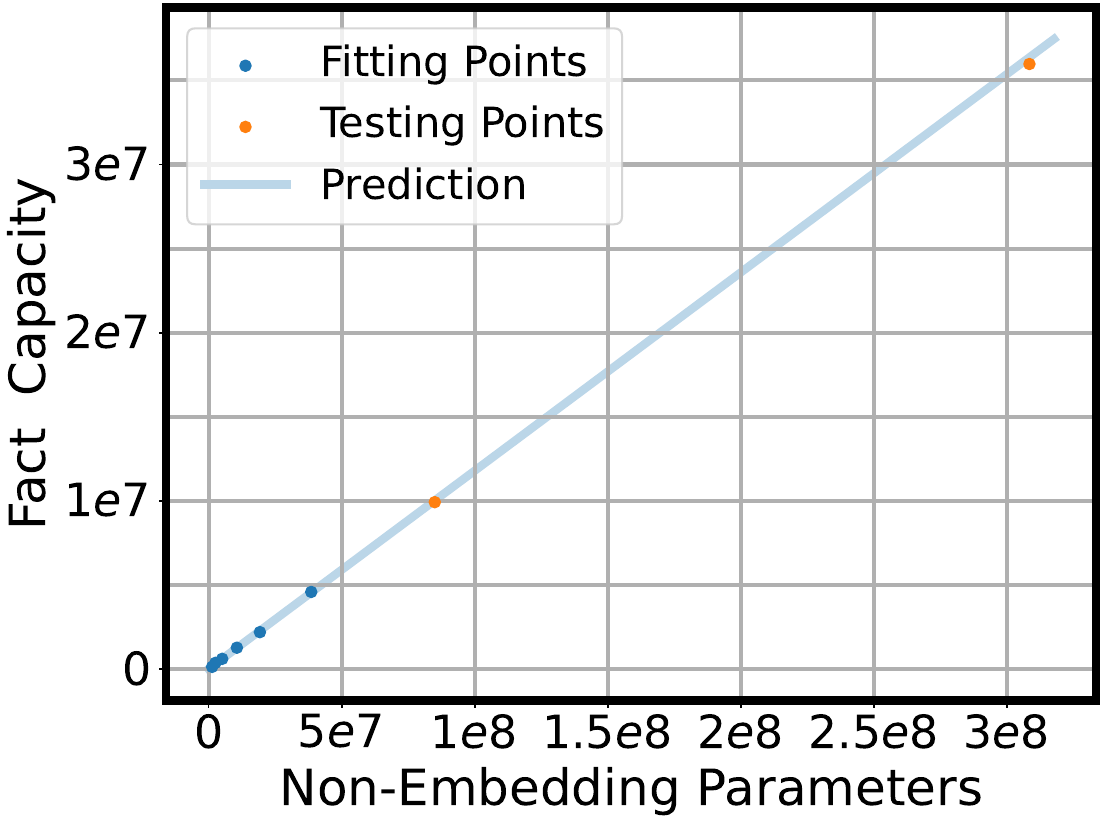}}
    \vspace{-5pt}
    \caption{The relation between LLMs' fact capacity and their model sizes, under fixed training epochs.}
    \label{fig:fact_capacity_and_model_size}
    \vspace{-10pt}
\end{figure}
We plot the fact capacities of the varying model sizes from 30M to 0.5B, under the fixed training epochs in Figure~\ref{fig:fact_capacity_and_model_size} (20M fails to reach 95\% MR at these epochs). 
We find that the LLM's fact capacity linearly scales with the model size. Meanwhile, we find that the line fitted from points of small model sizes (non-embed parameters $<=38\text{M}$ ) can extrapolate well to large model size 0.5B (308M non-embed parameters $\approx 8\times 38\text{M}$ ), which shows the robustness of the linear scaling laws.
\paragraph{Scaling Law of Fact Capacity and Epochs}
\begin{figure}[]
    \centering
    \subfigure[44M Model]{
    \includegraphics[width=0.22\textwidth]{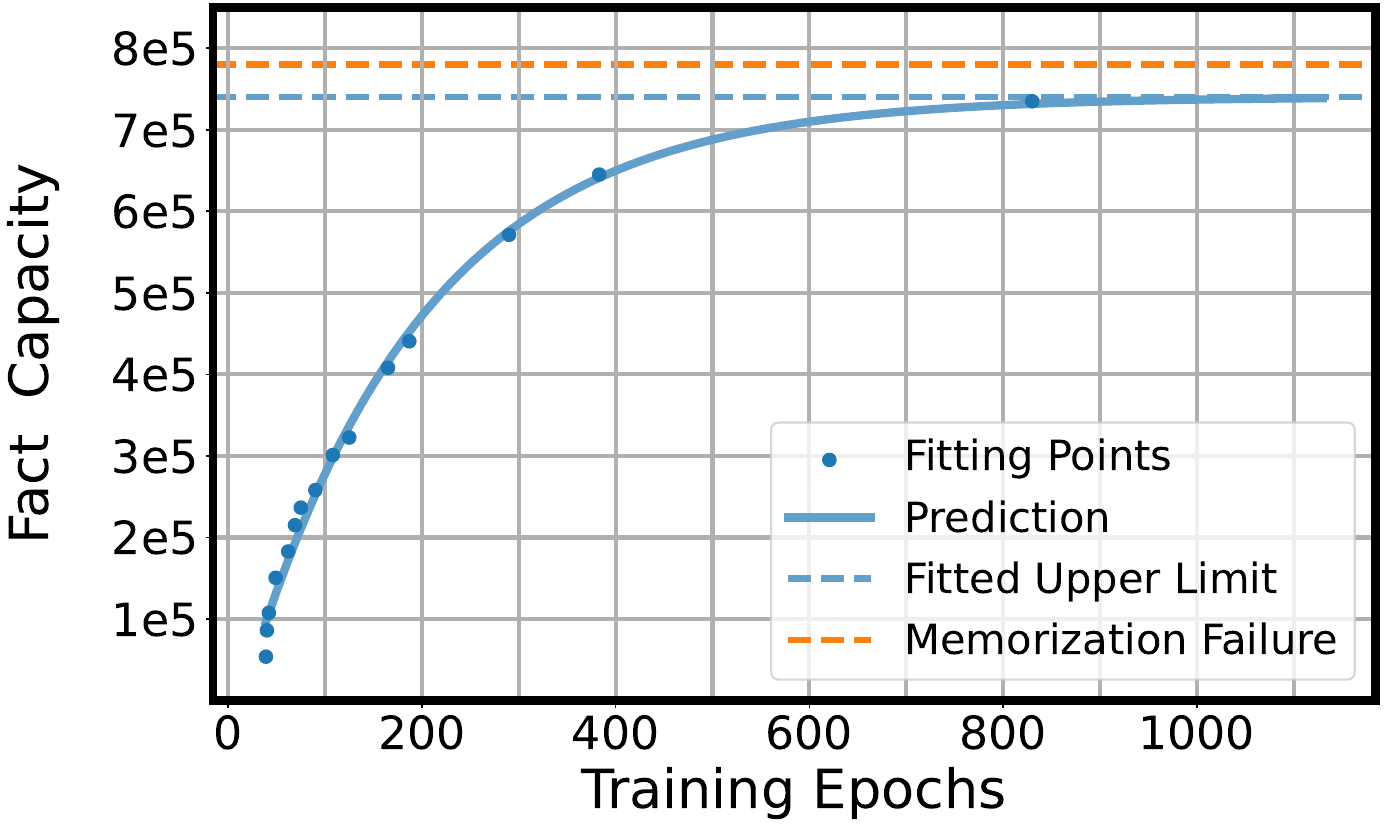}}
    \subfigure[69M Model]{
    \includegraphics[width=0.22\textwidth]{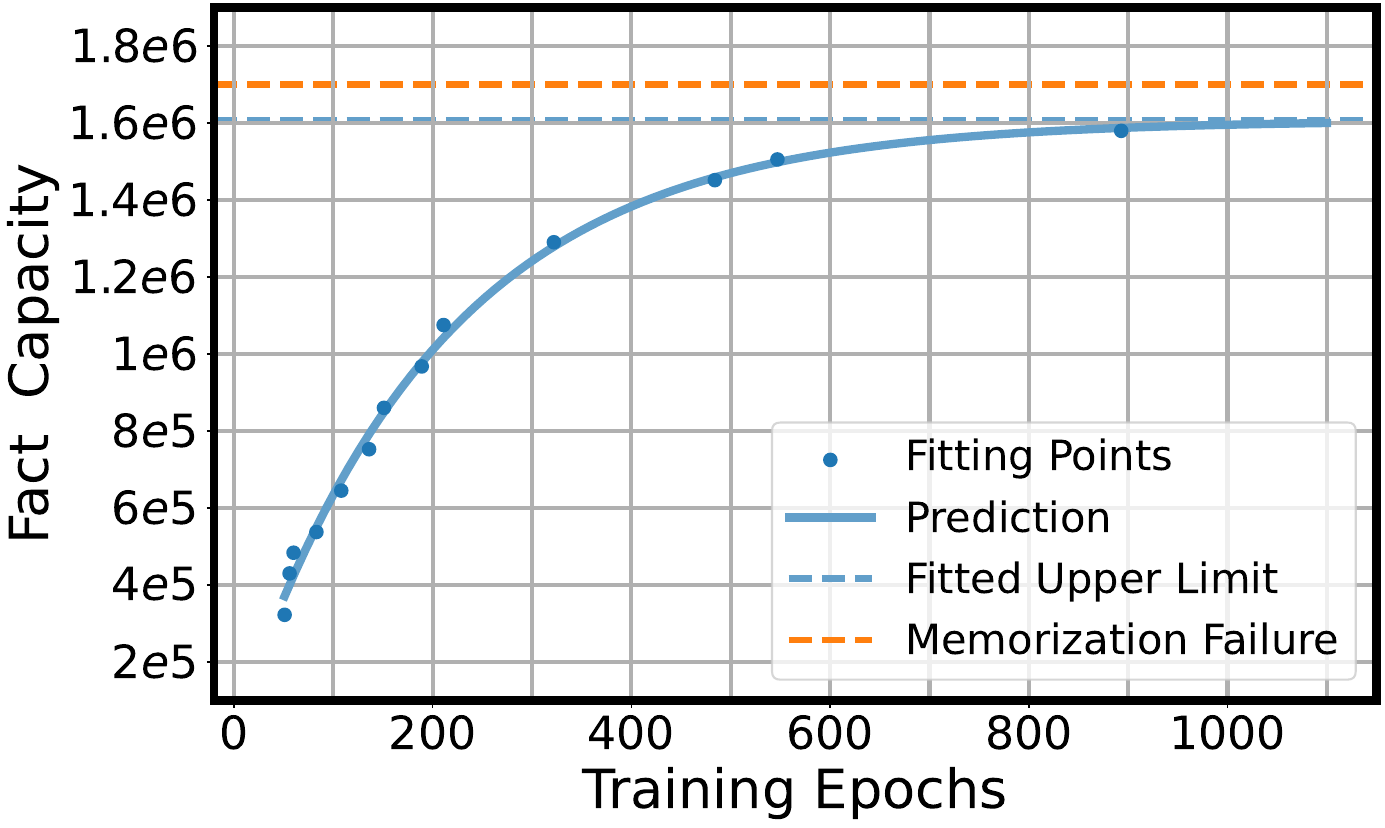}}
    \vspace{-8pt}
    \caption{The relation between LLMs' fact capacity and training epochs, under fixed model size.}
    \label{fig:fact_capacity_and_epoch}
    \vspace{-10pt}
\end{figure}
We plot the same LLM's fact capacities under varying training epochs in Figure~\ref{fig:fact_capacity_and_epoch}.
We find that with increasing training epochs, the LLM's fact capacity significantly increases at the beginning and then approaches saturation at about 1000 epochs, and we use the negative exponential law to fit the trend:
\vspace{-5pt}
\begin{align}
    C = C^* - \alpha_E \cdot \operatorname{exp} (-\beta_E \cdot Epoch),
\end{align}
where $C^*$ means the LLM's fact capacity saturation when epochs approach infinity, and $\alpha_E$ and $\beta_E$ are constants.
We further train the LLM on fact quantity which is 1.1 times of $C^*$ and then find the LLM fails to accurately memorize all of those training facts, under 3000 epochs (almost saturated), which verifies the effectiveness of the fitting of negative exponential law.
Additionally, for those small training epochs, e.g., $<35$, the LLM almost can not accurately memorize facts, and this shows 
that the cost of fact memorization is significantly higher than general knowledge learning by pre-training, which usually requires only one epoch~\citep{internlm2}.
This result indicates that it is challenging for the LLM to memorize those low-frequency fact knowledge in pre-training and up-sampling those facts can be a potential solution. 
\paragraph{Experiments on Wikidata}
We extend our experiments to Wikidata. 
Specifically, we use the fact triples from Wikidata as the training facts and plot the relation between the fact capacity and LLM's size in Figure~\ref{fig:fact_capacity_and_parameter_Wikidata}. We find that the results on Wikidata also show a linear relation between the fact capacity and the model size, which demonstrates the generality of the linear scale of the capacity parameter. According to the fitted line, we estimate that it requires an LLM with 1000B non-embed parameters to fully memorize all of Wikidata fact triples (about 15B\footnote{\href{https://www.Wikidata.org/wiki/Property:P10209}{https://www.Wikidata.org/wiki/Property:P10209}}) under 100 training epochs, which seems costly. Since Wikidata's fact knowledge is only a subset of all public facts, our analysis indicates that it is very challenging for an LLM to memorize all public fact knowledge in the common LLM size and pre-training setting, which shows the necessity of enhancing LLMs' fact knowledge by external information, e.g., RAG~\citep{rag_survey}.

\begin{figure}[]
    \centering
    \subfigure[Company $\to$ Credit-No]{
    \includegraphics[width=0.22\textwidth]{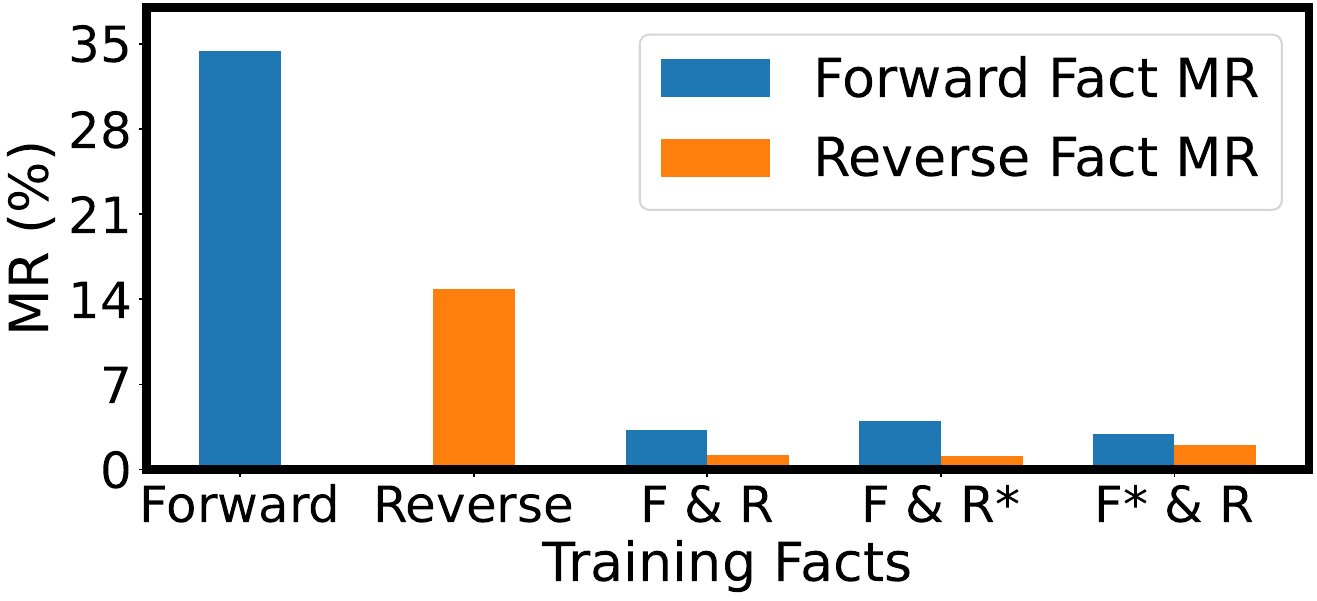}
    \includegraphics[width=0.22\textwidth]{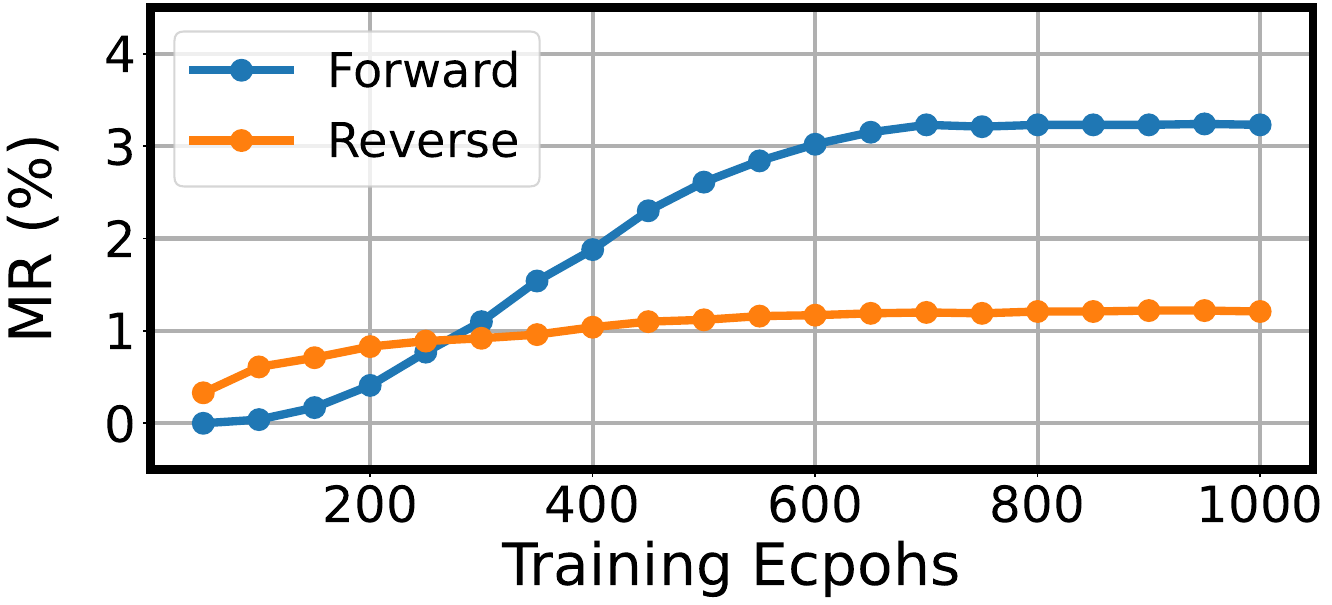}
    }\vspace{-5pt}
    
    \subfigure[Company $\to$ Operator]{
    \includegraphics[width=0.22\textwidth]{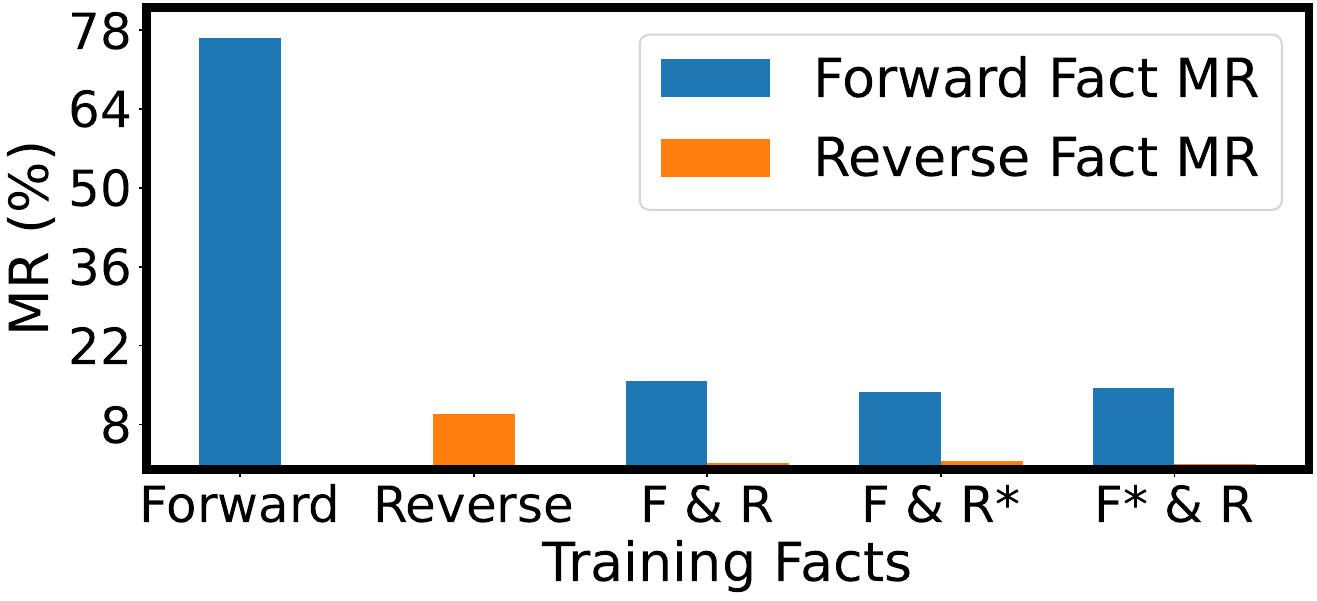}
    \includegraphics[width=0.22\textwidth]{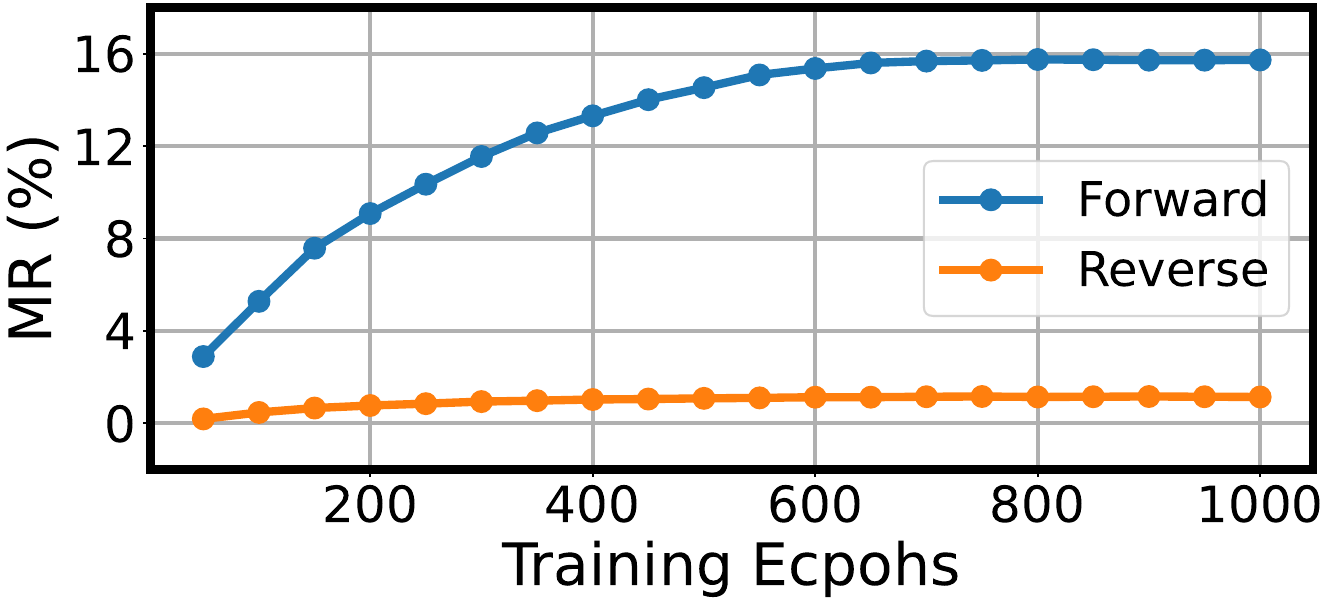}
    }\vspace{-5pt}
    \subfigure[Company $\to$ Register-No]{
    \includegraphics[width=0.22\textwidth]{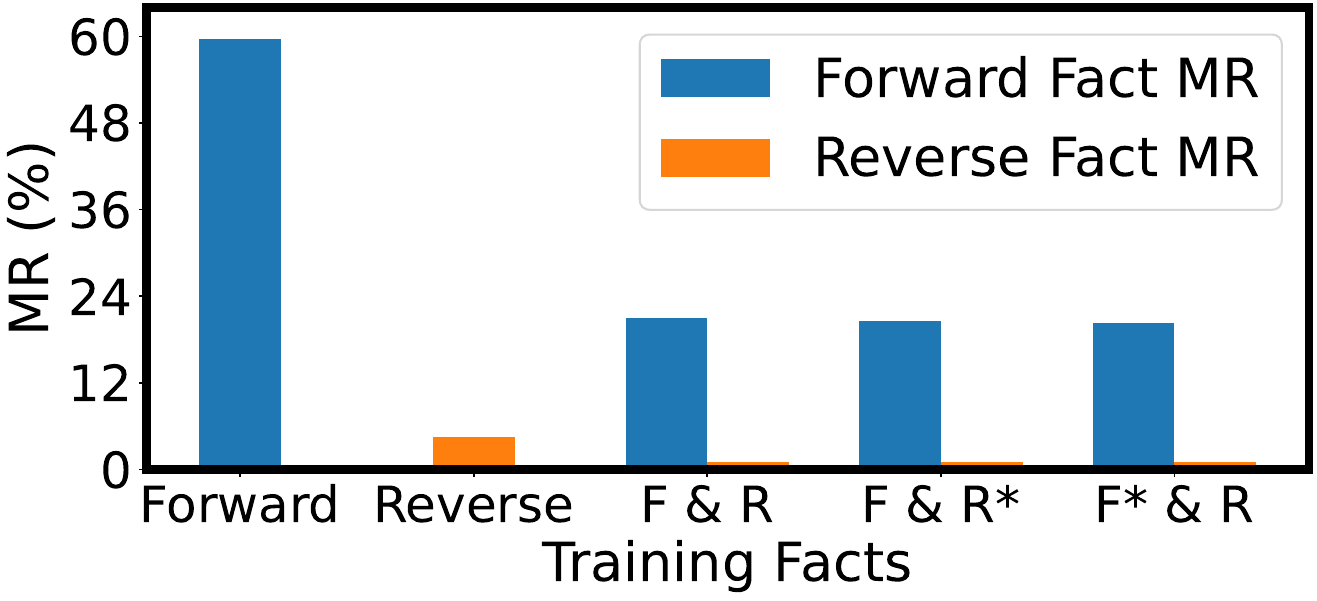}
    \includegraphics[width=0.22\textwidth]{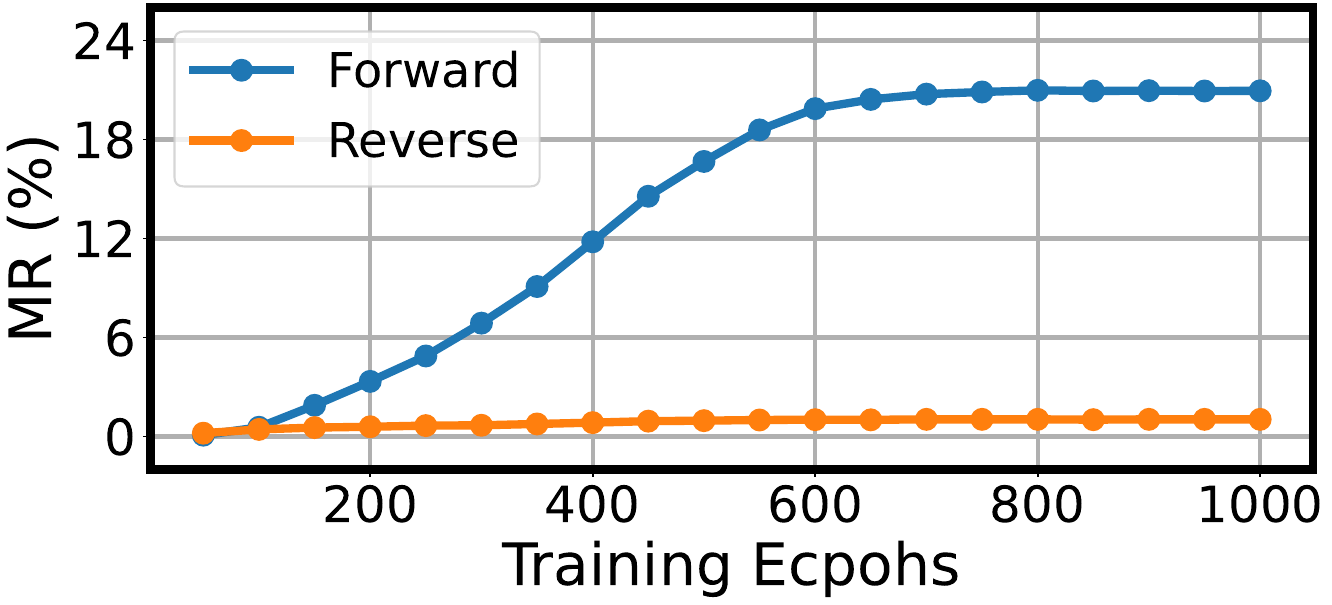}
    }
    \vspace{-10pt}
    \caption{LLMs' memorization of the same facts with different directions, 
    where ``*'' means facts are from another group of keys. 
    The right is the learning curves.
    }
    \label{fig:fact_compatibility_different_direction}
    \vspace{-15pt}
\end{figure}
\section{Redundant Fact Memorization}
\vspace{-5pt}
In this section, we explore whether LLMs can efficiently memorize redundant facts, i.e., whether LLMs can save memorization capacity when simultaneously memorizing redundant facts. Specifically, we conduct experiments on three types of redundant facts: 1) The forward and reverse versions of the same fact knowledge
2) The correlated facts of the same key 
3) Single-hop facts and their derivable multi-hop facts. Additionally, we analyze whether learning abstract abilities occupies the fact memorization capacity. We set the training epoch as 1000 to make the LLM's memorization saturated, 
unless otherwise specified.

\paragraph{The Same Fact of Different Directions}

In this section, we analyze whether the LLM can efficiently memorize the forward and reverse versions of the same facts. The forward fact is predicting the value based on the company name and attribute, as in Eq~\eqref{eq:value_prediction}. The reverse fact is predicting the company based on the attribute's value~\citep{reverse_curse, llm_physics_32}.
We select three highly reversible attributes, ``Operator'', ``Credit-No'' and ``Register-No''for the experiment.
Specifically, we compare the memorization rate of the following three groups: 1) separately memorizing the forward or reverse version of the same facts. 2) simultaneously memorizing the forward and reverse versions of the same facts (redundant). 3) simultaneously memorizing the forward facts and the reverse version of another set of facts (non-redundant). The number of each direction's facts is the same and thus the memorization load of group 2 and 3 is consistent. We show the results on 41M model in Figure~\ref{fig:fact_compatibility_different_direction}. We also plot corresponding learning curves in Figure~\ref{fig:fact_compatibility_different_direction}, which show that the LLMs' fact memorization is almost saturated.
The results on 30M model are shown in Appendix~\ref{app:fact_memorization_different_direction_30m} and show similar trends.
We see that simultaneously memorizing facts of different directions leads to a significantly lower MR than separately memorizing them and the MR of simultaneous memorization is lower than the half of separate memorization.

These show that the LLM does not compatibly memorize them and memorizing different directions of the same fact even conflicts with each other.
Meanwhile, memorizing different directions of the same group of facts (redundant) has a similar memorization rate to memorizing different groups of facts (non-redundant). These show that when the LLM memorizes the same facts in different directions, it seems to memorize them separately like memorizing independent facts, which reflects the inefficiency of LLM memorization for the same facts with different directions~\citep{reverse_training}.
Since the massive facts can be described in different directions, these results can indicate that the LLM's parametric knowledge is not efficient for fact memorization.

\paragraph{Correlated Facts of the Same Key}
\begin{figure}[]
    \centering
    \includegraphics[width=0.22\textwidth]{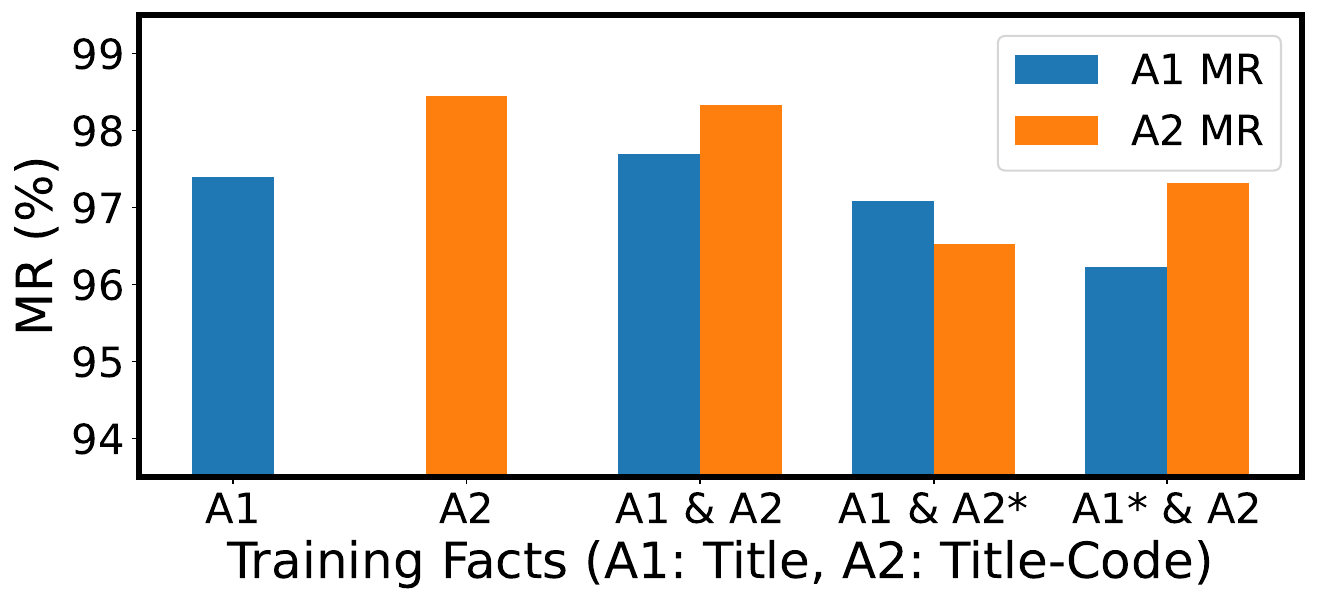}
    \includegraphics[width=0.22\textwidth]{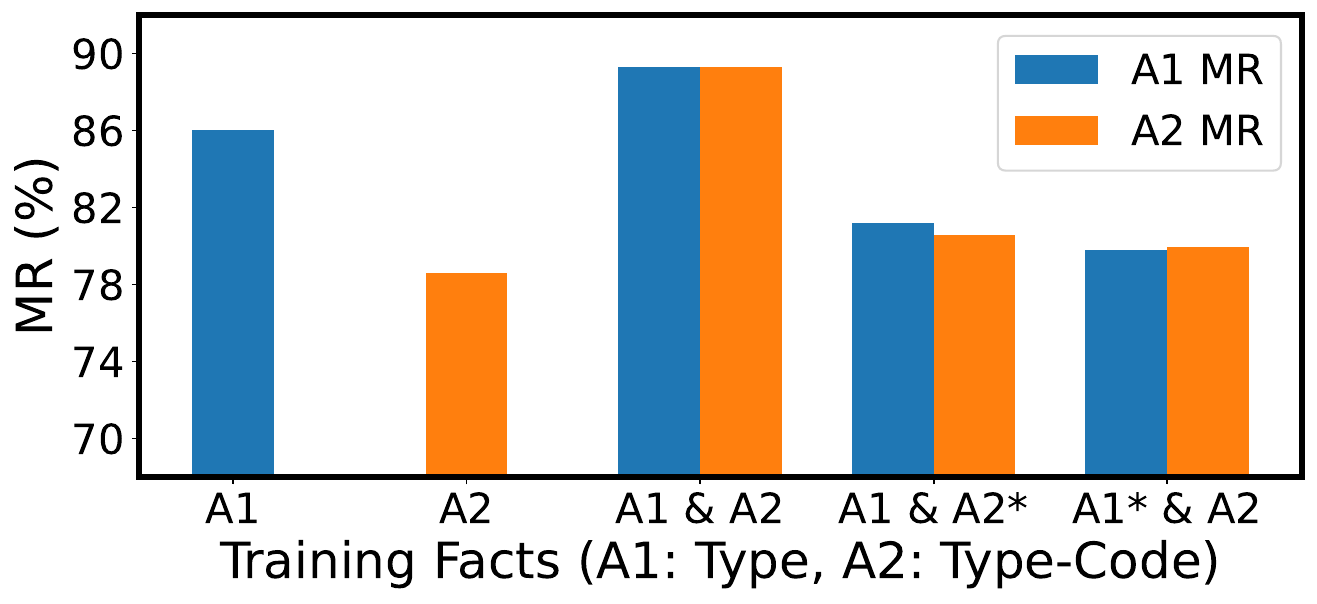}
    
    \includegraphics[width=0.22\textwidth]{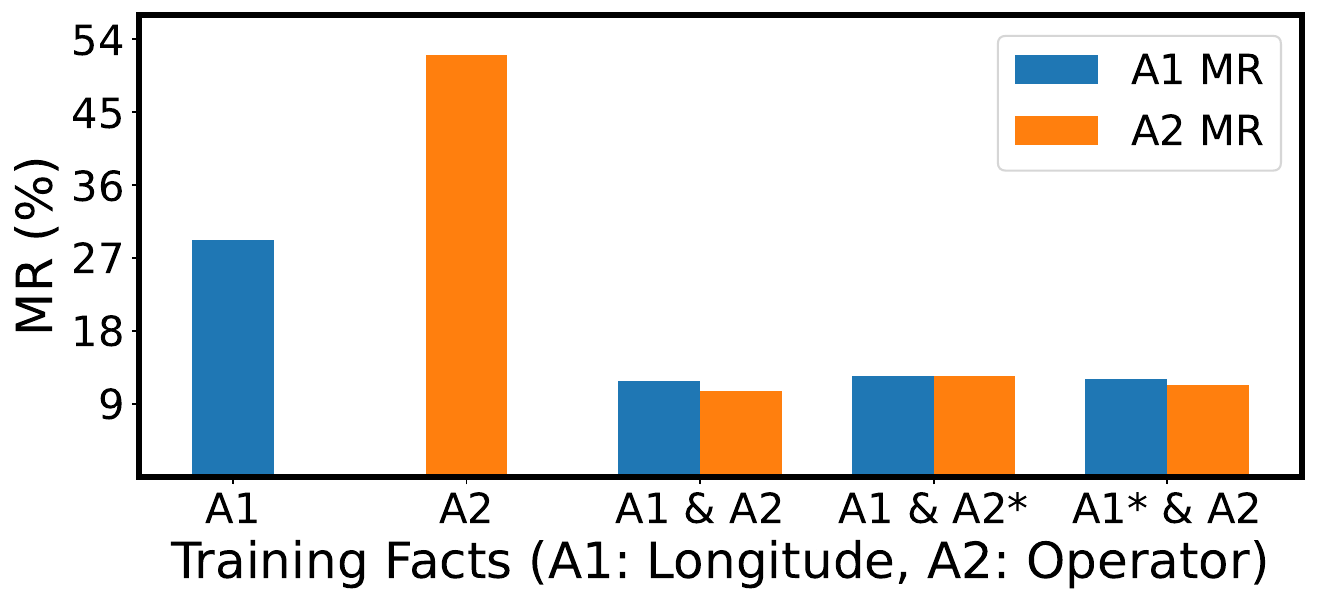}
    \includegraphics[width=0.22\textwidth]{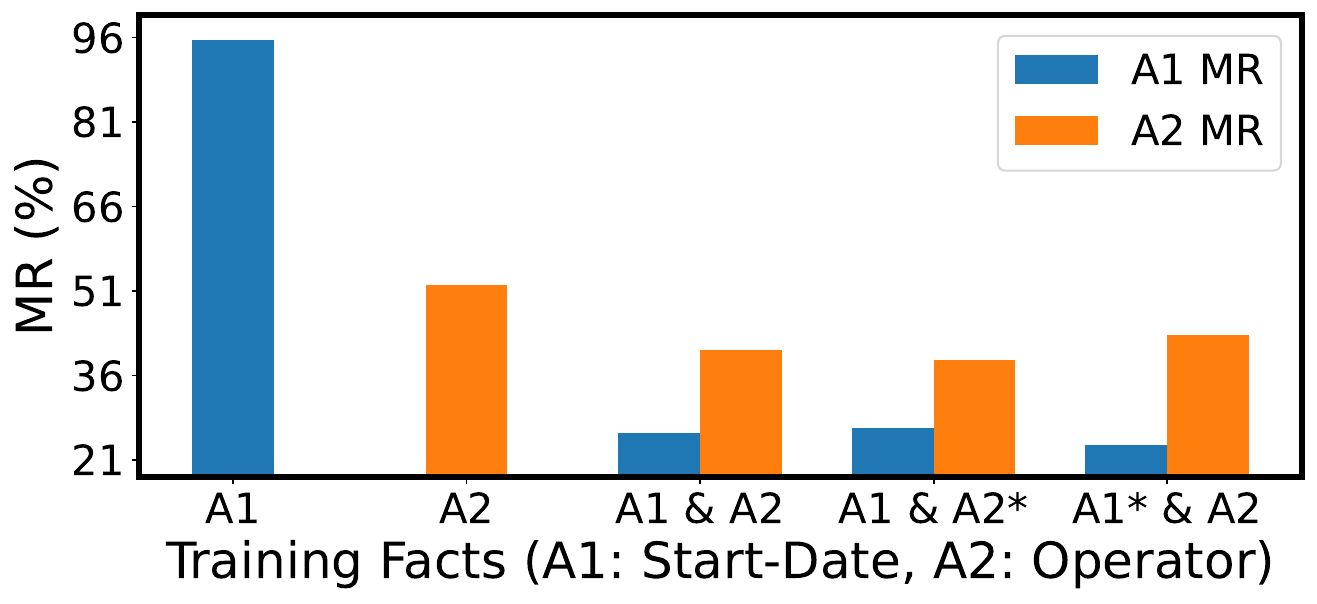}

    \vspace{-5pt}
    \caption{Memorization on correlated facts, where ``*'' means that facts are from another group of keys.}
    \label{fig:fact_compatibility_different_attribute}
    \vspace{-15pt}
\end{figure}

In this section, we analyze whether the LLM can efficiently memorize the correlated facts of the same key, e.g., a company's type and its type code. Specifically, we select two combinations of correlated attributes to conduct analysis and additionally adopt two unrelated combinations as a comparison. For each combination, we compare the memorization rate of the following three groups: 1) individually memorizing facts of a single attribute; 2) simultaneously memorizing facts of two attributes on the same companies (if attributes are correlated, these facts will be redundant);
3) simultaneously memorizing one attribute's facts on a group of companies and another attribute's facts on another group of companies (non-redundant). The number of each attribute's facts is the same and thus the memorization load of group 2 and 3 is consistent.

The results are shown in Figure~\ref{fig:fact_compatibility_different_attribute}. 
We find that simultaneously memorizing correlated attributes leads to a higher memorization rate than separate memorization, which shows LLMs can efficiently memorize one key's correlated attributes, and correlated fact memorization can facilitate the individual fact's memorization. Meanwhile, for those unrelated attributes, simultaneously memorizing them leads to a decreased memorization rate, which shows that whether LLM can compatibly memorize one key's facts highly depends on the correlation of those facts.
While it is hard to inject new correlated knowledge into LLMs~\citep{llm_physics_32}, these results indicate the potential of additionally memorizing correlated facts in pre-training since they can be compatibly memorized.

\begin{figure}[]
    \centering
    \subfigure[Longitude]{
    \includegraphics[width=0.22\textwidth]{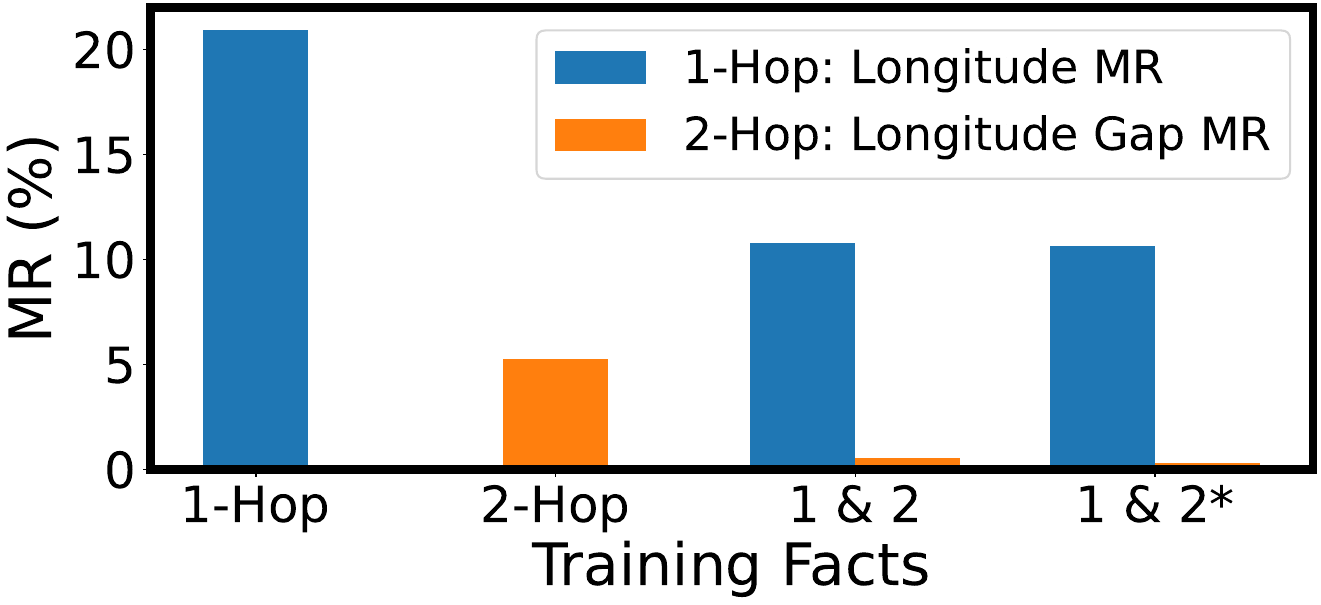}}
    \subfigure[Start-Date]{
    \includegraphics[width=0.22\textwidth]{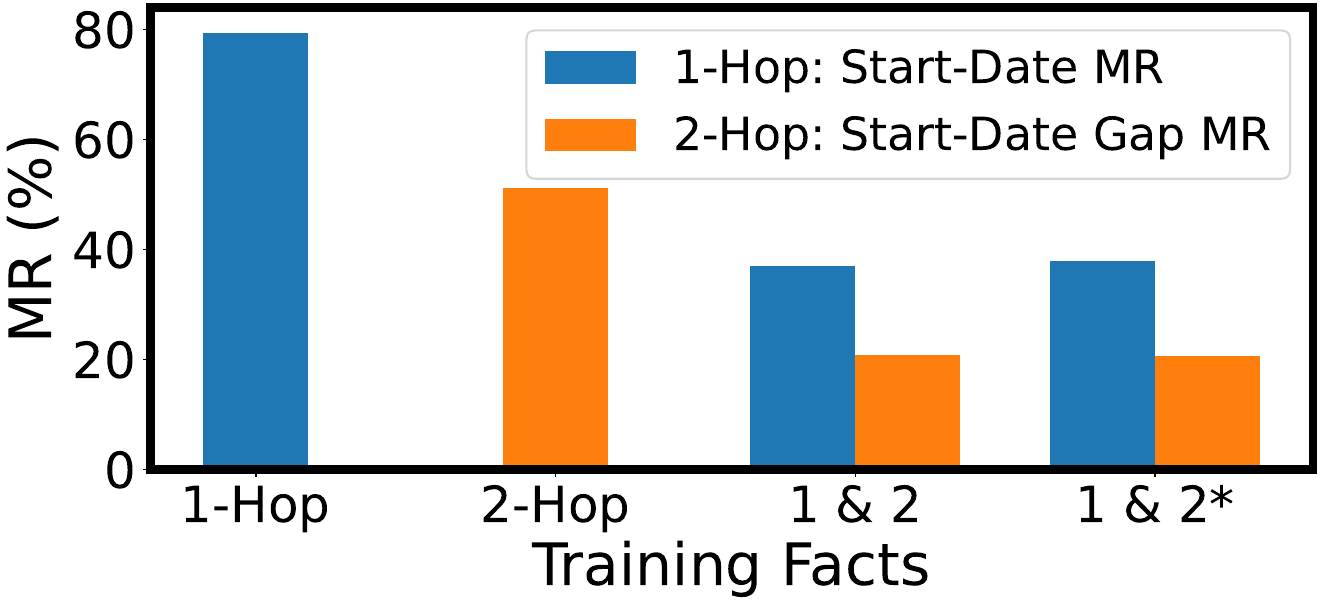}}
    
    \vspace{-10pt}
    \caption{LLM memorization for derivable facts, where ``*'' means that facts are from another group of keys.}
    \vspace{-10pt}
    \label{fig:fact_memorization_compatibility_derivable_fact}
\end{figure}

\paragraph{Derivable Multi-hop Fact}
In this section, we analyze whether the LLM can efficiently memorize derivable facts. For example, when the LLM memorizes the longitude of two companies, can it additionally memorize their longitude gap efficiently? We explore this question on facts about attributes ``Longitude'' and ``Start-Date'', and choose their gap as derivable 2-hop facts. 
For 2-hop facts of one attribute, given two different keys, we train the LLM to predict the value gap of this attribute.
Specifically, we compare the memorization rate of the following three groups: 1) separately memorizing single-hop facts and their derivable 2-hop facts. 2) simultaneously memorizing single-hop facts and their derivable 2-hop facts (redundant). 3) simultaneously memorizing single-hop facts and 2-hop facts derived from another set of single-hop facts (non-redundant). 
We control the numbers of 1-hop facts and 2-hop facts to be equal.
The results are shown in Figure~\ref{fig:fact_memorization_compatibility_derivable_fact}. We find that group 2 leads to a significantly lower memorization rate than group 1, which shows that the memorization of derivable 2-hop facts is not compatible with corresponding 1-hop facts.
Additionally, the memorization rate of group 2 is similar to group 3. This shows that when the LLM memorizes single-hop facts and their derivable facts, it seems to memorize them separately like memorizing irrelevant facts. This reflects the inefficiency of LLM memorization for derivable facts, which hinders the LLM's fact capacity for massive derivable facts in pre-training corpus~\citep{multihop_fact_shortcut}. 

\begin{figure}[]
    \centering
    \subfigure[Longitude \& SNLI]{
    \includegraphics[width=0.22\textwidth]{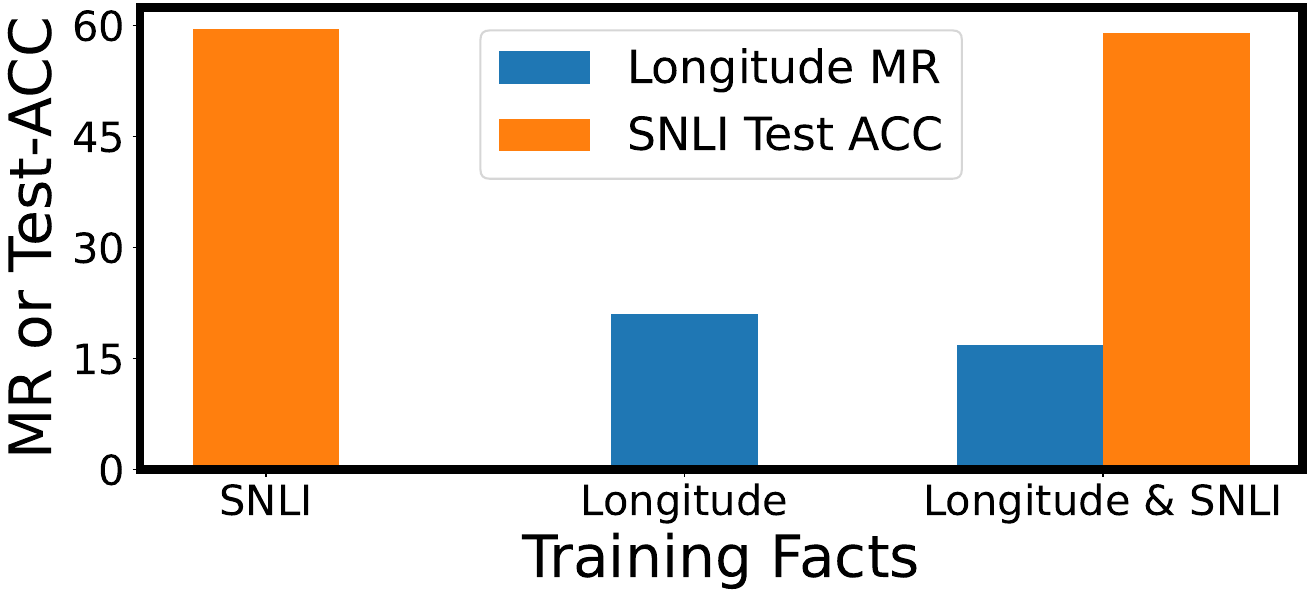}}
    \subfigure[Longitude\&Amazon-CLS]{
    \includegraphics[width=0.23\textwidth]{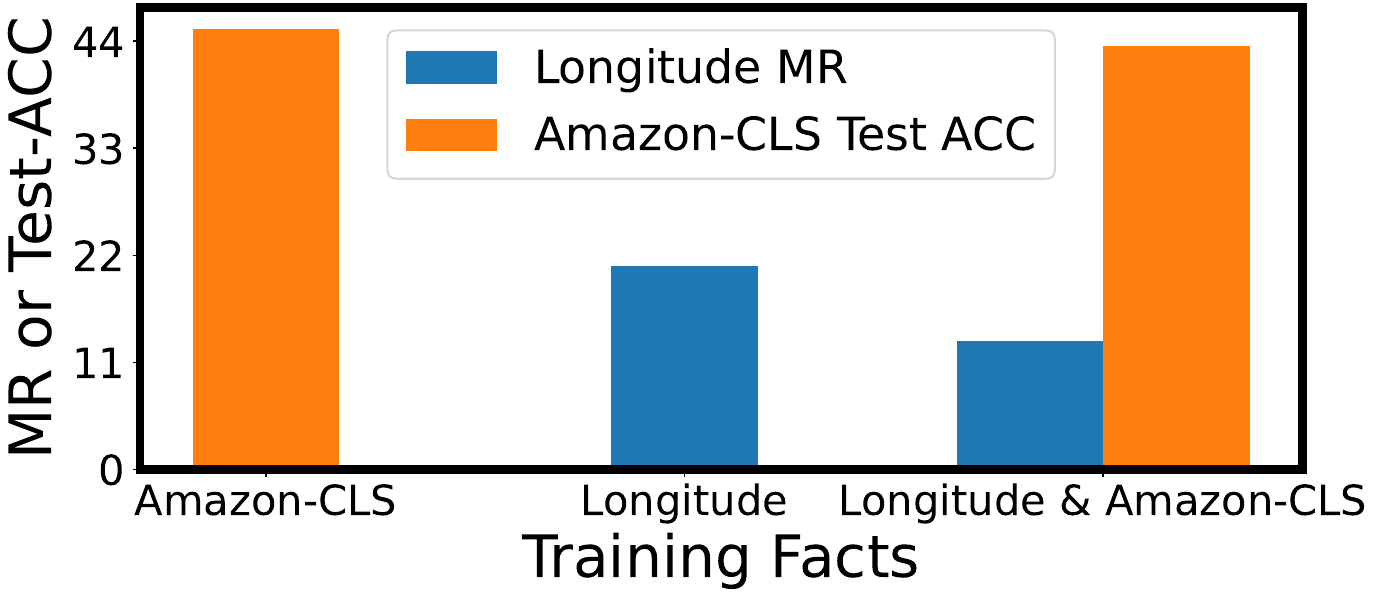}}
    
    \vspace{-8pt}
    \caption{The influence of abstract ability learning to LLM's fact memorization.}
    \vspace{-10pt}
    \label{fig:fact_memorization_compatibility_abstact_ability}
\end{figure}
\paragraph{Fact Memorization Meets Abstract Ability Learning}

We explore whether abstract ability learning occupies LLMs' fact memorization capacity. Specifically, we compare the fact MR or test accuracy of two groups: 
1) separately learning fact knowledge and abstract ability
2) simultaneously learning fact knowledge and abstract abilities. We use SNLI~\citep{dataset_snli} and Amazon Sentiment Analysis~\citep{dataset_amazon} for abstract ability learning. The frequency of facts and abstract ability examples is the same.
The results are shown in Figure~\ref{fig:fact_memorization_compatibility_abstact_ability}. We see that additionally learning abstract abilities decreases the fact memorization rates. Meanwhile, the incorporation of fact knowledge slightly hurts the classification tasks' test accuracy. These indicate that fact memorization and abstract ability learning will influence each other and occupy the LLMs' knowledge capacity jointly, which further exacerbates the challenges of LLMs memorizing facts during pre-training.

\vspace{-3pt}
\section{Fact Memorization Preference of LLMs}
\vspace{-4pt}
We analyze LLMs' fact memorization preference in three aspects: frequency, difficulty and memorization order. Since this section focuses on preference, we select irrelevant facts to conduct experiments. Specifically, we use the combination of facts in the company information table and a specialized subset of Wikidata facts (Book $\to$ Author).
\paragraph{Frequency}
\begin{figure}[t]
    \centering
    \subfigure[Longitude \& Author]{
    \includegraphics[width=0.22\textwidth]{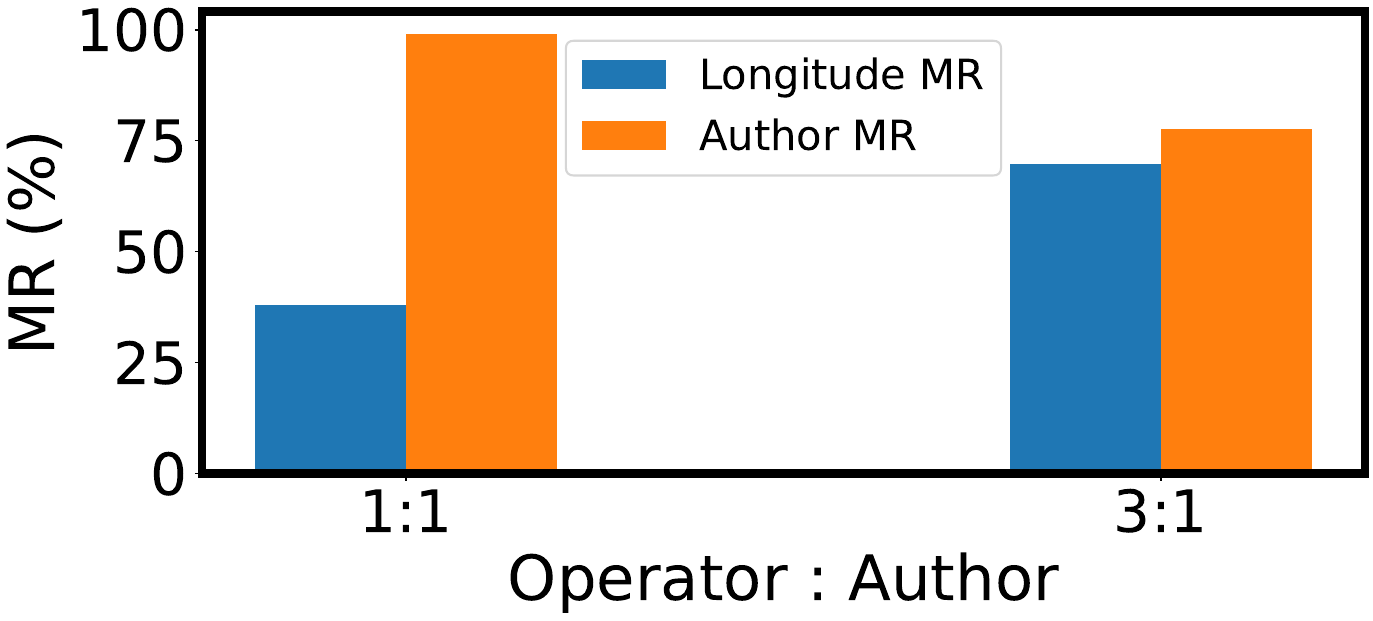}}
    \subfigure[Operator \& Author]{
    \includegraphics[width=0.22\textwidth]{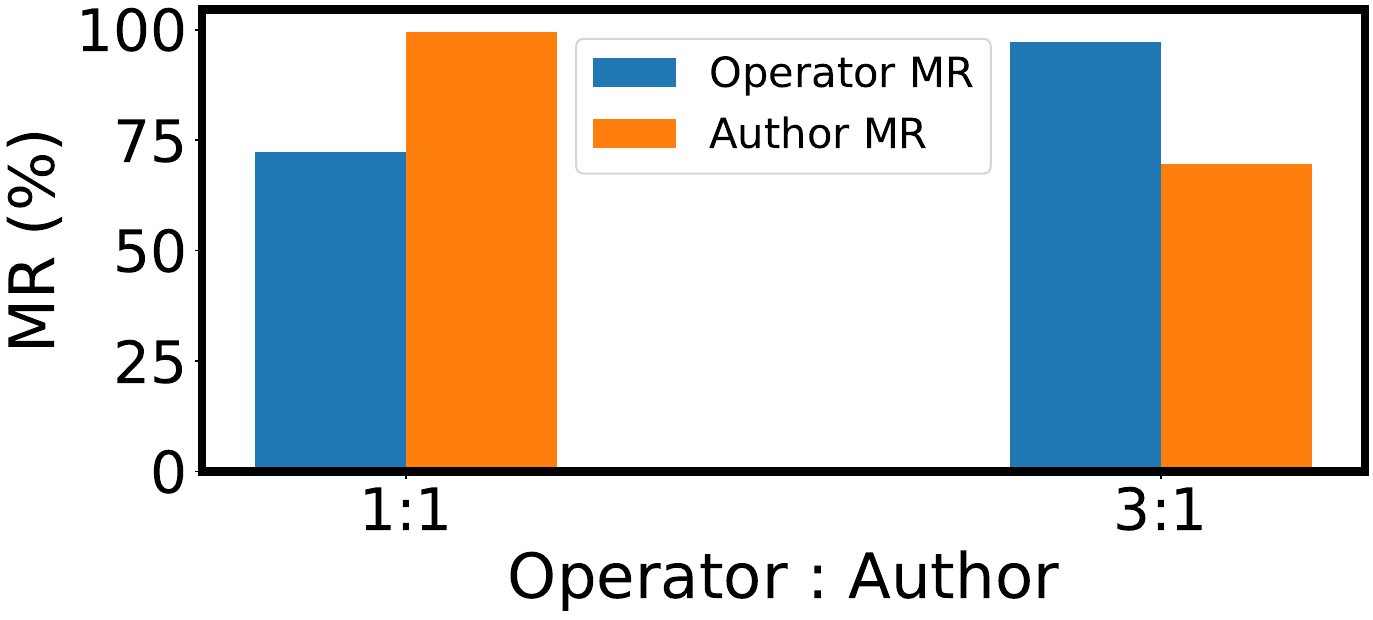}}
    
    \vspace{-10pt}
    \caption{The effect of frequency for fact memorization.}
    \label{fig:fact_memorization_preference_frequency}
    \vspace{-10pt}
\end{figure}
We compare the respective memorization rate of simultaneously memorizing two attributes under different frequencies. The results on ``Longitude \& Author'' and ``Operator \& Author'' are shown in Figure~\ref{fig:fact_memorization_preference_frequency}. We see that the higher frequency leads to a significantly higher memorization rate and inhibits low-frequency facts' memorization~\citep{popqa}. 
This indicates the importance of increasing the frequency of low-frequency facts in pre-training corpus to facilitate LLMs' memorization of them. However, since facts in pre-training corpus usually appear in a complicated and mixed manner, it is non-trivial to separately control their respective frequency, which further increases the challenges for LLMs to memorize low-frequency facts.

\begin{figure}[]
    \centering
    \subfigure[Longitude \& Credit-No]{
    \includegraphics[width=0.22\textwidth]{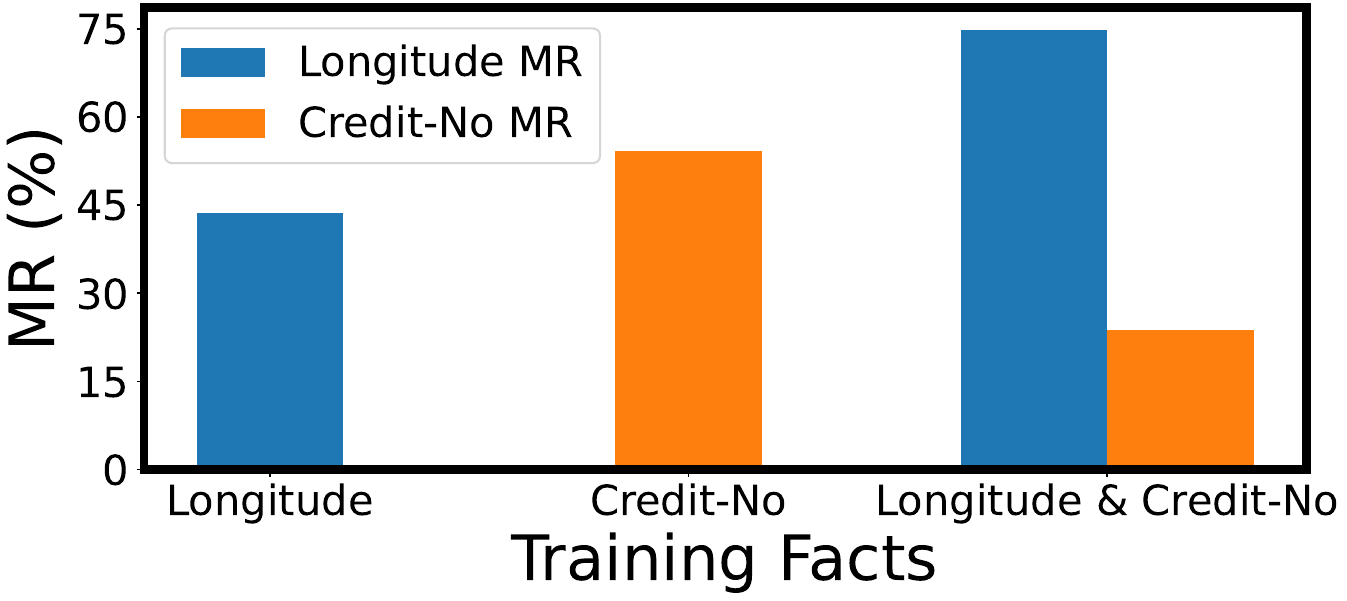}}
    \subfigure[Longitude \& Operator]{
    \includegraphics[width=0.22\textwidth]{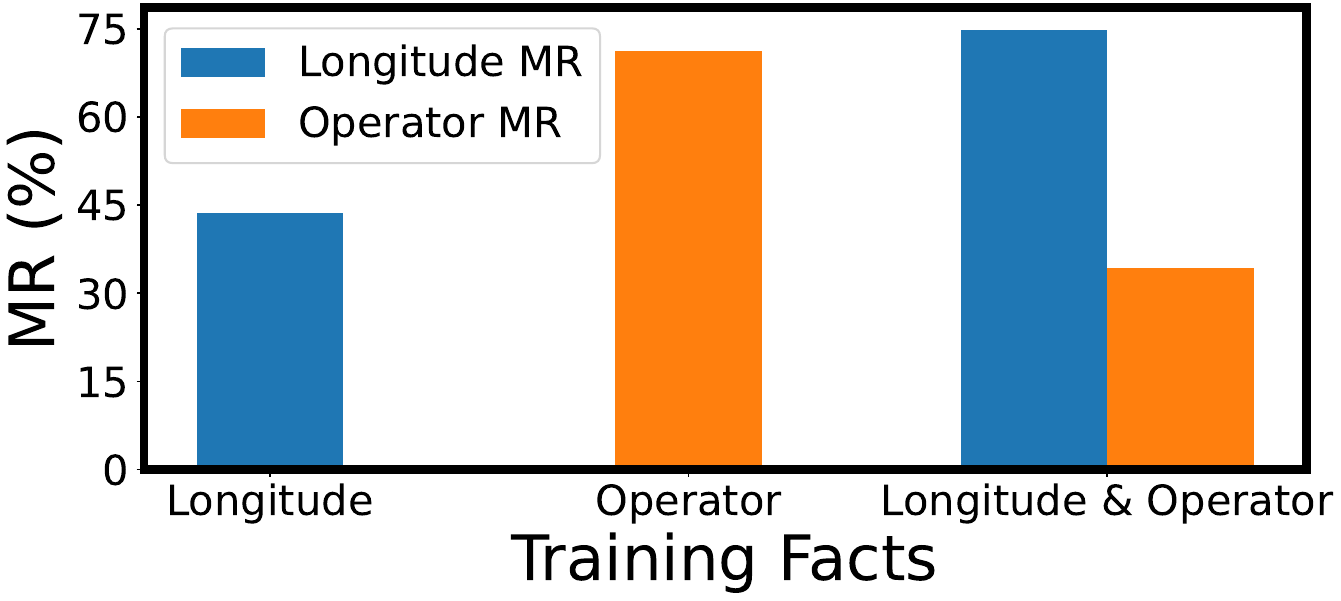}}
    
    \vspace{-10pt}
    \caption{The effect of difficulty for fact memorization.}
    \label{fig:fact_memorization_preference_difficulty}
    \vspace{-4pt}
\end{figure}
\vspace{-5pt}
\paragraph{Difficulty}
We compare the respective memorization rate of three groups: 
1) using LLM of size $2*N$ to simultaneously memorize facts of two attributes with different memorization difficulties; 2) using LLM of size $N$ to separately memorize facts of each attribute.
In group 1 and 2, the number of facts of each attribute is the same and thus the average fact capacity for each attribute is consistent. In this way, we can observe the LLM's preference when simultaneously memorizing two attributes.
The results on ``Longitude \& Credit-No'' and ``Longitude \& Operator'' are shown in Figure~\ref{fig:fact_memorization_preference_difficulty}. In group 2, the memorization rates of attributes ``Credit-No'' and ``Operator'' are lower and thus they are harder to memorize. Compared with group 2, the memorization rate of difficult facts and easy facts in group 1 increases and decreases, respectively. 
These show that when LLMs memorize different types of facts, they tend to pay more attention to the facts that are harder to memorize.
\vspace{-5pt}
\paragraph{Memorization Order}
\begin{table}[]
\small
\centering
\begin{tabularx}{\columnwidth}{@{}l>{\hsize=1.2\hsize\centering\arraybackslash}X>{\hsize=0.8\hsize\centering\arraybackslash}X@{}}
\toprule
\textbf{Training Facts}   & \textbf{Longitude MR} & \textbf{Author MR} \\ \midrule
\textbf{Longitude}        & 20.9                  & -                  \\
\textbf{Author}           & -                     & 76.1               \\
\textbf{Longitude$\Rightarrow$Author} & 0          & 13.1               \\
\textbf{Author$\Rightarrow$Longitude} & 17.7       & 0                  \\ \bottomrule
\end{tabularx}

\vspace{1.5mm}

\begin{tabularx}{\columnwidth}{@{}l>{\hsize=1.0\hsize\centering\arraybackslash}X>{\hsize=1.0\hsize\centering\arraybackslash}X@{}} 
\toprule
\textbf{Training Facts}     & \textbf{Credit-No MR} & \textbf{Operator MR} \\ \midrule
\textbf{Credit-No}          & 30.7                  & -                    \\
\textbf{Operator}           & -                     & 38.9                 \\
\textbf{Credit-No$\Rightarrow$Operator} & 0         & 32.6                 \\
\textbf{Operator$\Rightarrow$Credit-No} & 20.6      & 0.1                  \\ \bottomrule
\end{tabularx}
\vspace{-5pt}
\caption{The influence of memorization order. ``A$\Rightarrow$B'' means memorizing A before B.}
\label{fig:fact_memorization_preference_order}
\vspace{-10pt}
\end{table}

\begin{figure}[]
\vspace{-10pt}
    \centering
    \subfigure[44M]{
    \includegraphics[width=0.22\textwidth]{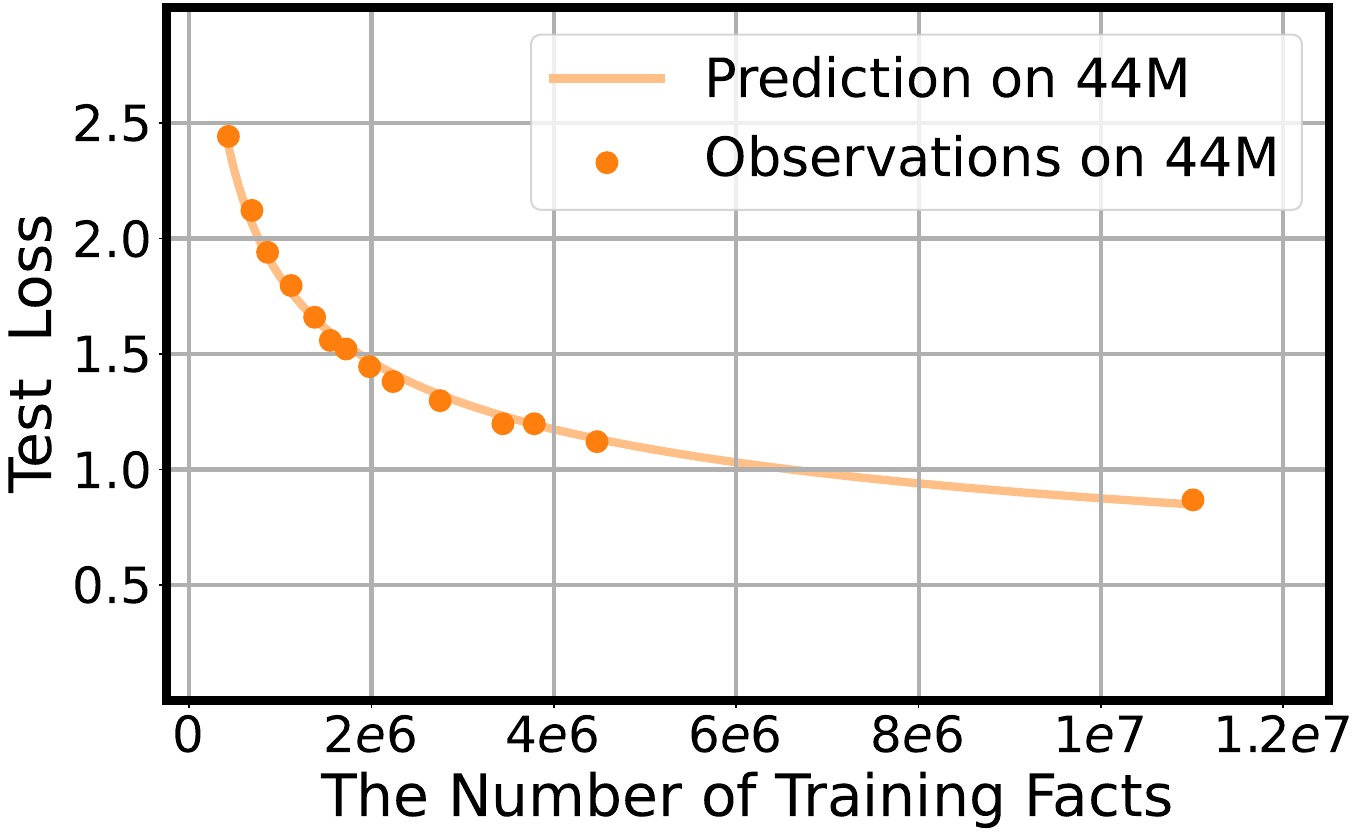}}
    \subfigure[97M]{
    \includegraphics[width=0.22\textwidth]{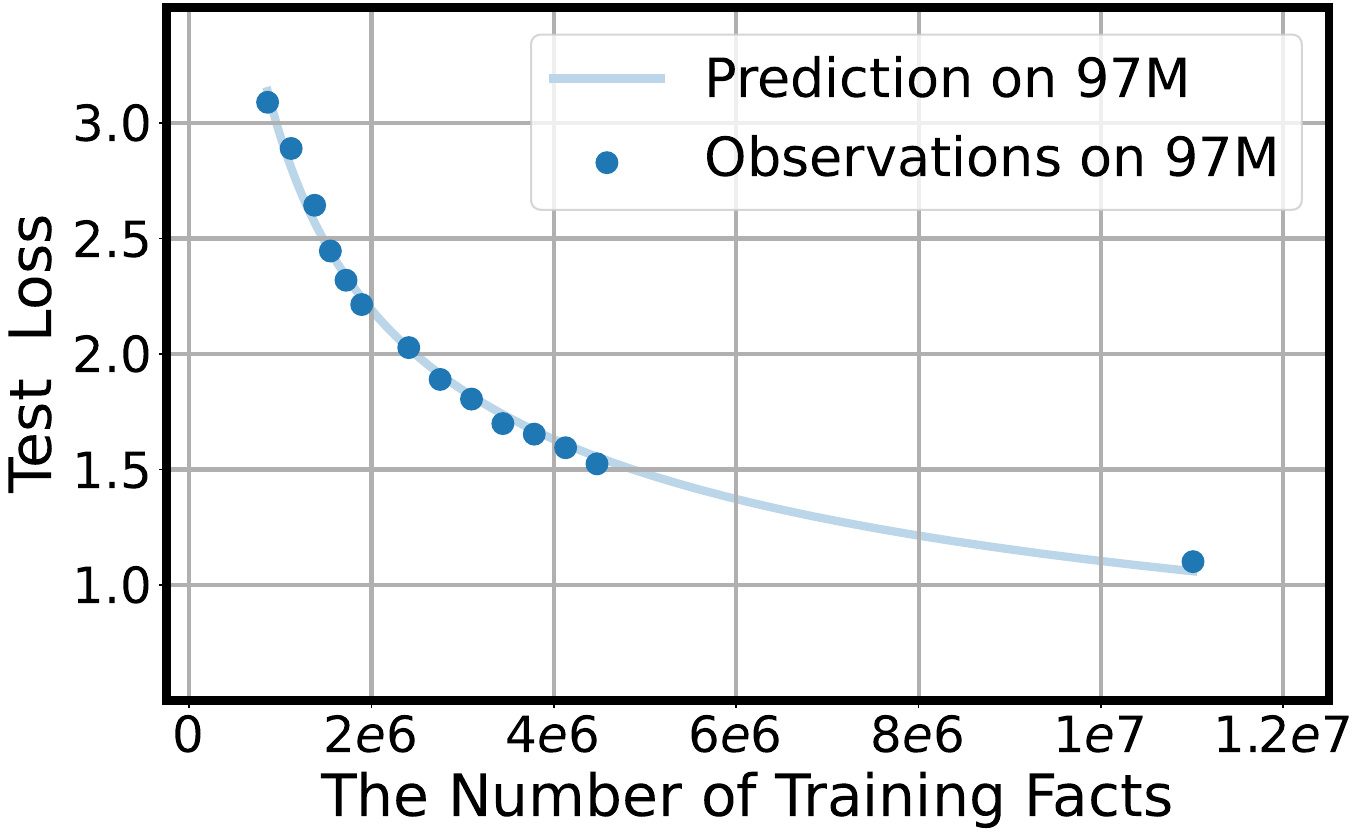}}
    \vspace{-10pt}
    \caption{LLMs' fact generalization loss across different numbers of training facts.}
    \label{fig:fact_generalization_scaling_law}
        \vspace{-15pt}
\end{figure}
\begin{figure*}[t]
    \centering
    \includegraphics[width=0.7\textwidth]{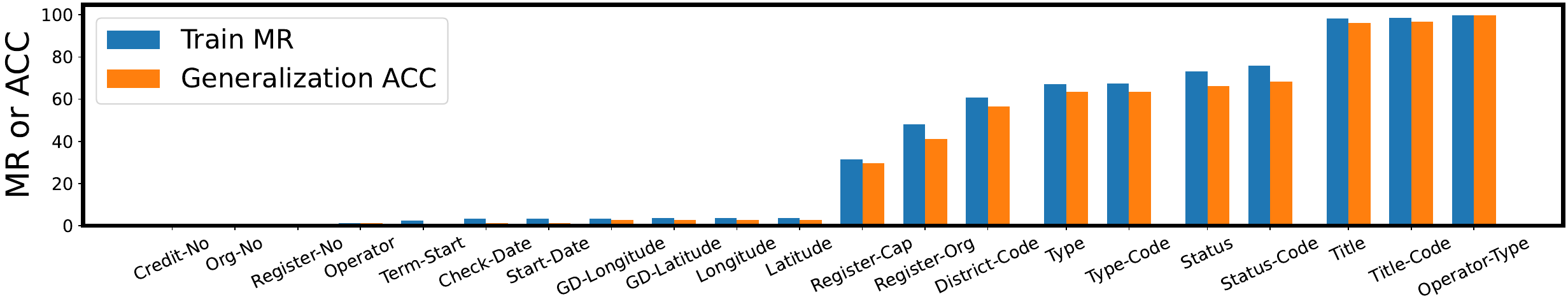}
    \vspace{-10pt}
    \caption{Memorization and generalization over facts of different types, on 44M LLM trained by 10M facts.}
    \vspace{-15pt}
    \label{fig:mr_test_acc_relation_44M}
\end{figure*}
We compare the memorization rate of simultaneously memorizing facts of two attributes in different memorization orders. The results are shown in Table~\ref{fig:fact_memorization_preference_order}. We find that the memorization rate of earlier facts decreases to almost zero and 
the subsequent memorized facts almost refresh the LLM's fact memorization. 
These indicate a potential reason for LLM's inferior memorization of low-frequency facts: maybe some of them only appear in the early stage of pre-training process and they were almost overwritten by the subsequent pre-training knowledge. Additionally, the MR of subsequent facts is lower than its individual memorization, which demonstrates the importance of evenly distributing various types of facts in pre-training process.
\vspace{-5pt}
\section{Fact Generalization of LLMs}
\vspace{-5pt}
Beyond the fact memorization, we explore an interesting question: can LLMs generalize on unseen fact knowledge? 
Specifically, we train the LLM to memorize facts of a group of keys and test it on unseen keys' facts.
We show each attribute's generalization accuracy (exact match) of 44M model
in Figure~\ref{fig:mr_test_acc_relation_44M}.
We also test results of 30M model and observe similar trends (see Appendix~\ref{app:memorization_and_generalization_30m}).
We observe that facts on most of the attributes have a generalization accuracy greater than zero, which indicates that LLMs can generalize on unseen fact knowledge to a certain level. 

To analyze why the LLM can generalize on fact knowledge, we conduct a case study on facts of three attributes and show the cases in Appendix~\ref{app:generalization_case_studies}. 
We find that LLMs' fact generalization depends on the correlation between input (key) and output (value)~\citep{short_cut_learning}. 
For a specific type of fact (attribute), the higher correlation between the key and value leads to higher generalization accuracy. 
For example, the LLM may correctly predict an unseen company's longitude if the company name contains a region name and the training dataset contains the longitude of companies with the same region name.
Or it can roughly estimate the company's register-capital according to company size indicated by the company name, e.g., ``Fruit shop''$\to$ (\textyen $10^4 \sim$ \textyen $10^5$) or ``Investment company''$\to$ (\textyen $10^7 \sim$ \textyen $10^9$). 
Meanwhile, different types of facts have different generalizability. For those facts with obvious patterns, the LLM can achieve reliable generalization.
For those attributes with weak correlation, although the LLM does not know exactly the facts, it can identify the rough range of facts.
These suggest the potential of adaptively leveraging LLM's fact generalization: 1. selectively leveraging generalization of those highly generalizable facts;
2. if the LLM does not exactly know the whole fact, it can response with a part of the fact, e.g., a rough range, to make its response more informative and thus helpful.
\paragraph{Fact Generalization Scaling Law}

Additionally, we analyze the scaling law of LLMs' fact knowledge generalization. Specifically, we plot the LLMs' loss values on test fact knowledge under different training fact quantities, following~\citet{scaling_law_for_lm}. The results are shown in Figure~\ref{fig:fact_generalization_scaling_law}.
We find that the test loss on fact generalization also follows the power-law~\citep{scaling_law_for_lm} as:
\vspace{-2pt}
\begin{align}
    L(D) = D_c * D^{\alpha_D},
\end{align}
where $D$ is the number of training facts, $D_c$ and $\alpha_D$ are constant numbers. 
This trend is similar with general pre-training~\citep{scaling_law_for_lm}, 
which indicates that LLMs follow a similar learning mechanism in learning factual knowledge as they learn general knowledge in pre-training~\citep{gpt4}.

\paragraph{Relation between Fact Memorization and Generalization}
We plot the memorization rate and generalization accuracy for each type of fact in Figure~\ref{fig:mr_test_acc_relation_44M}. We find that the generalization accuracy of one type of fact highly correlates with its memorization rate. For one type of fact, the higher memorization rate leads to higher generalization accuracy. 
These indicate that both LLM fact memorization and generalization are based on the correlation between input and output~\citep{short_cut_learning}.
If there is a stronger correlation between the input and output, it will be easier for the LLM to memorize and learn about the type of fact knowledge in a unified manner. If the correlation is minimal, LLM needs to memorize facts individually, and is hard to generalize on unseen ones.
\vspace{-2pt}
\section{Related Work}
\vspace{-3pt}
Understanding the scaling behaviors of LLMs is important for decisions about the key choice design of LLMs, e.g., model size or pre-training data~\citep{scaling_law_for_lm, scaling_law_reward_model_overoptimization, scaling_law_moe}. Most of the existing work focuses on the scenario of general pre-training or downstream tasks. \citet{scaling_law_for_lm} observe the power law relationships between the LLM perplexity and size of LLM and dataset. 
\citet{scaling_law_compute_optimal_llm} explore the optimal token quantity and LLM size for pre-training under a specified compute budget and find that the LLM size and training tokens should be scaled equally for compute-optimal LLM training. 
Besides pre-training, researchers find that the performance of downstream tasks can be predicted from the LLM size and training data scale
\citep{scaling_law_for_transfer,scaling_law_translation, scaling_law_downstream}.
Different from them, our paper specifically focuses on scaling laws of LLMs' fact memorization and behaviors of memorizing different types of facts, which is critical for LLMs' factual responses.

\vspace{1pt}

\citet{llm_physics_33}, concurrently to our work, explore scaling laws of LLMs' memorization on synthetic facts. Our work differs in several ways: 1. We analyze LLM's fact memorization on real-world facts while they use randomly generated facts, which have a non-negligible gap with real-world facts. 
According to our findings, we conclude that memorizing all Wikidata's facts requires 1000B non-embed parameters, which indicates that using an LLM to memorize all public facts is almost not plausible.
2. We additionally analyze LLMs' behaviors of learning fact knowledge in different aspects, including compatibility, preference and generalization, which further provide directions for fact knowledge augmentation of LLMs.
\vspace{-10pt}
\section{Conclusion}
\vspace{-2pt}
We analyze LLMs' fact memorization behaviors and these are our main conclusions:
1) The fact capacity has a linear relationship with model size and a negative exponential law relationship with training epochs. 
According to the built scaling law, we 
estimate that memorizing all of Wikidata fact triples requires training an LLM with 1000B non-embed parameters for 100 epochs, which seems very costly; 2) We find that LLMs struggle with efficiently memorizing redundant facts. Only for redundant facts with the same direction and structure, LLMs can memorize them in a unified manner. 3) The LLM prefers memorizing more frequent and difficult facts. 4) LLMs can generalize on unseen fact knowledge and its scaling law is similar to general pre-training.
\section*{Limitations}
We list limitations of this paper as follows:
\begin{itemize}
    \item Since this paper focuses on fact knowledge memorization, each atomic fact individually forms a training example and we keep the same inputs for the training and inference stages. This has a small gap with pre-training setting, which usually uses unstructured text and concatenates short sentences into a large chunk for training efficiency. 
    We regard the exploration of facts of unstructured text as future work.
    \item As shown in Figure~\ref{fig:fact_capacity_and_epoch}, fact memorization requires hundreds of training epochs, which leads to significant computational costs. Limited by computational resources, the maximum LLM size used in experiments is 0.5B. We regard the exploration of larger scales as future work.
\end{itemize}
\section*{Ethics Statement}
In this paper, we use public fact information for experiments, including a real-world company information table and Wikidata fact triples. The company information table is provided by a commercial data company and we obtain its permission to conduct this research. Meanwhile, the trained models are only for LLM fact memorization analytical research and will not be made public.

\bibliography{custom}
\appendix
\clearpage
\section{The Effect of Template Quantity}
\label{app:template_quantity_effect}
\begin{figure}[!h]
    \centering
    \subfigure[Longitude]{
        \includegraphics[width=0.22\textwidth]{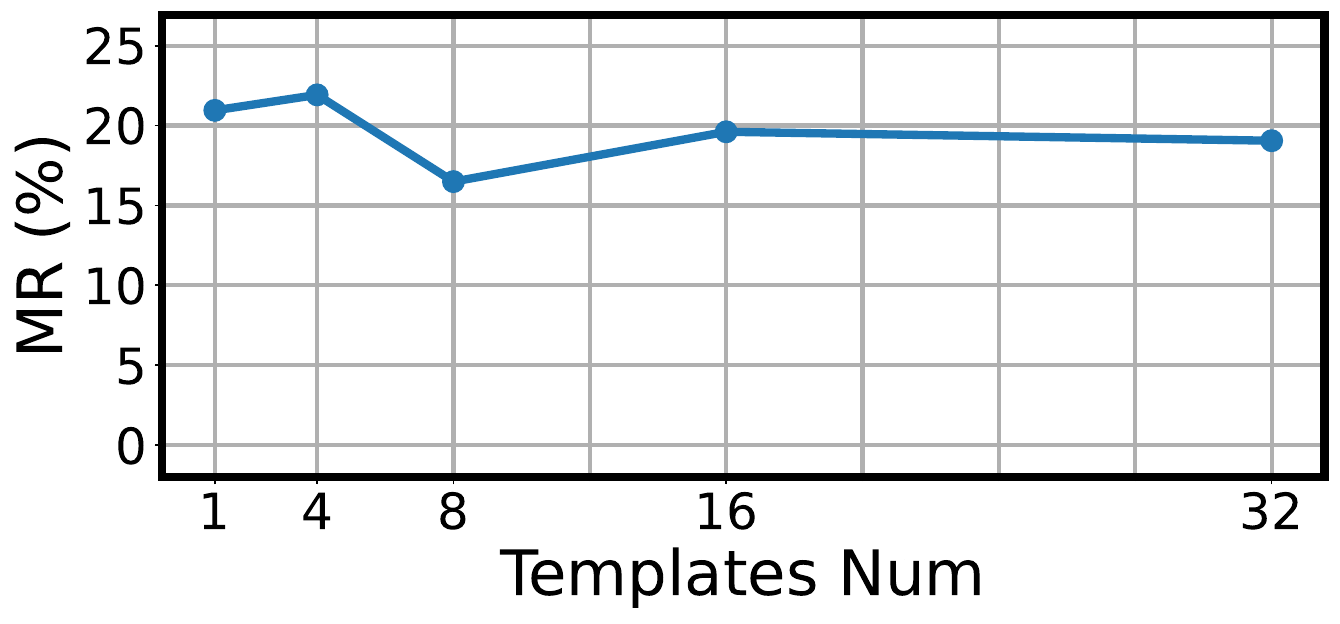}
    }
    \subfigure[Start-Date]{
        \includegraphics[width=0.22\textwidth]{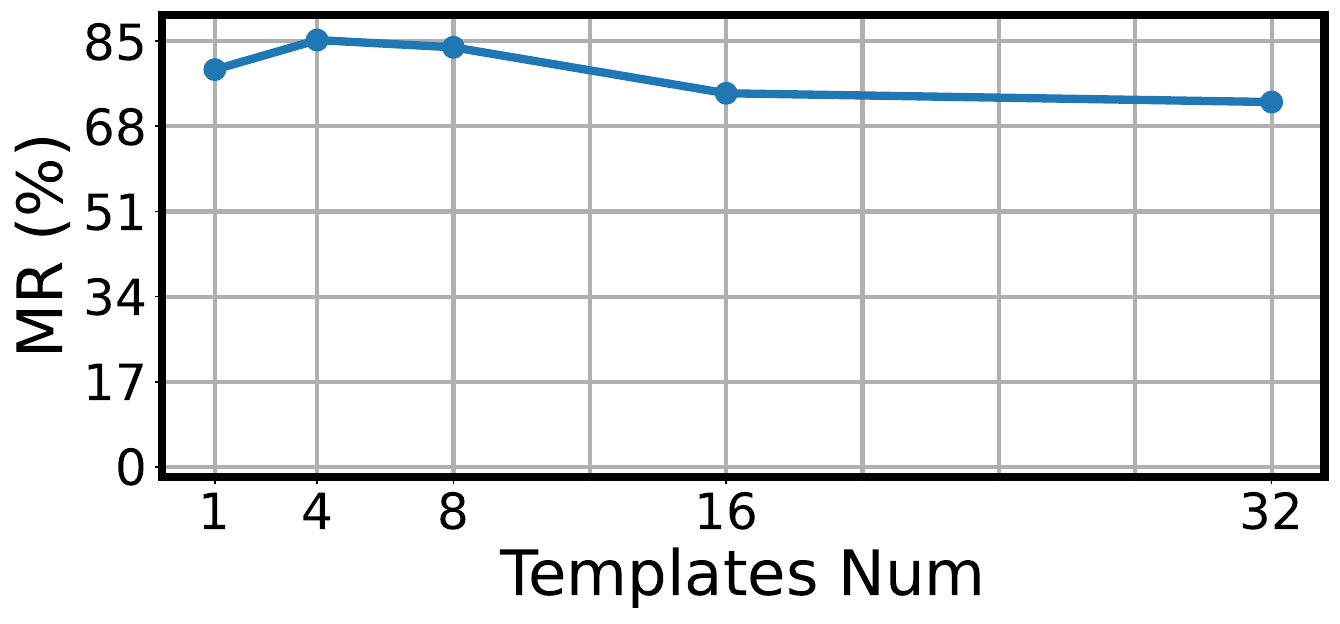}
    }
    \subfigure[Operator]{
        \includegraphics[width=0.22\textwidth]{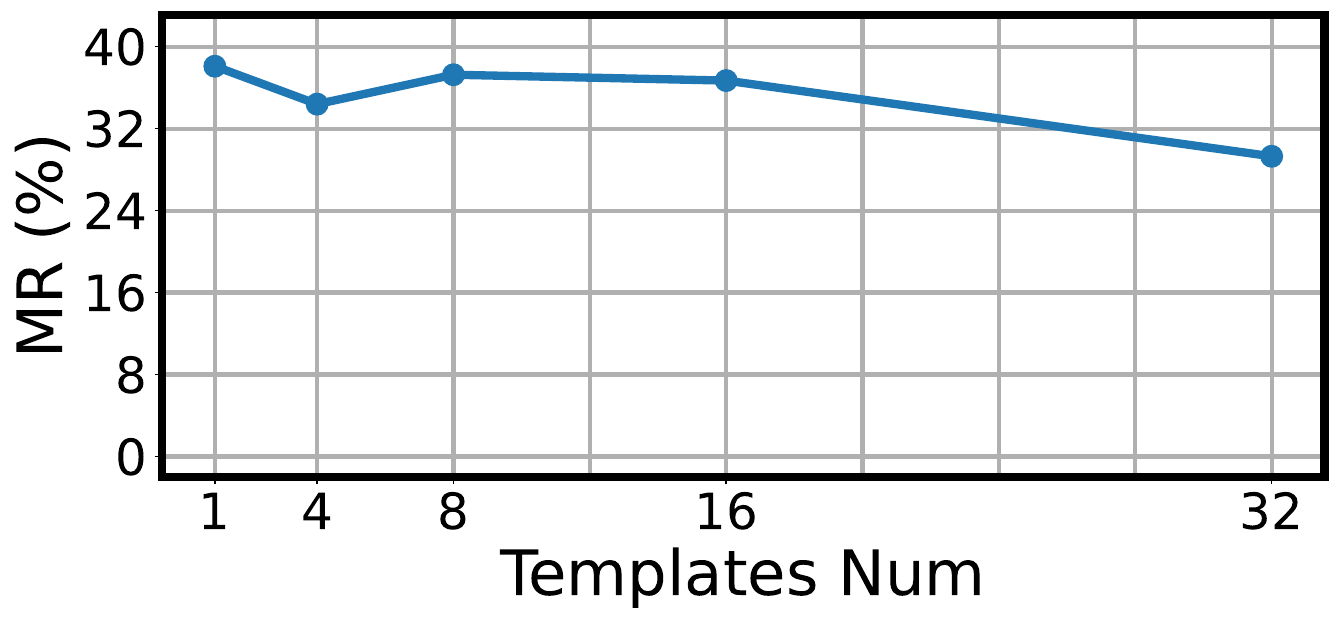}
    }
    \subfigure[Credit-No]{
        \includegraphics[width=0.22\textwidth]{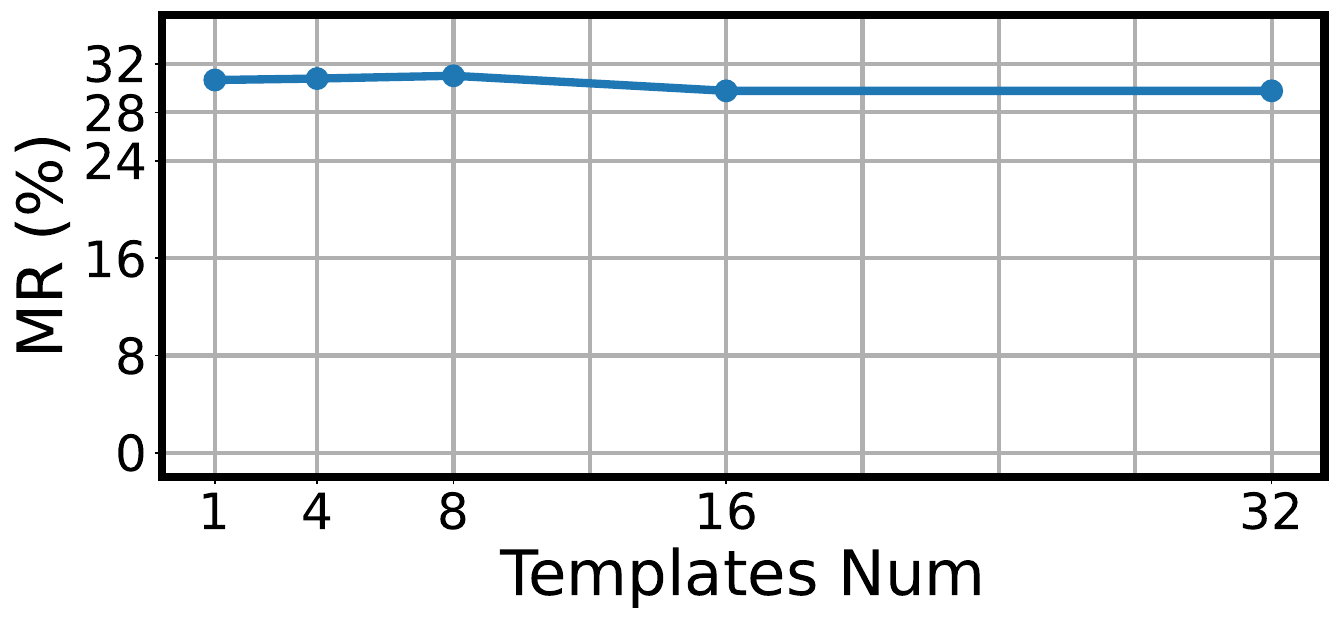}
    }
    \caption{Memorization on different templates.}
    \label{fig:fact_compatibility_paraphrase}
\end{figure}

In this section, we analyze the influence of template quantity on memorization rate. Specifically, we observe the memorization rate of the same 200K facts under various numbers of templates using a 30M model. The results are shown in 
Figure~\ref{fig:fact_compatibility_paraphrase}. 

We see that the memorization rates over different numbers of templates are at a consistent level. Even when the paraphrase quantity increases to 32, the memorization rate of specific attribute facts only decreases to as low as 75\% of the original. Therefore, the number of templates does not significantly influence the LLM's fact memorization.

\section{Attributes of Table}
\label{app:overall_table_information}
We list the information and average length of overall fields of the used large information table in Table~\ref{tab:table_all_fields}.

\section{The Effect of Pre-training on Fact Memorization}
\label{app:pre_training_effect}
We compare the fact knowledge learning from scratch and pre-trained checkpoints. The results are shown in Table~\ref{tab:pre_train_comparison}. We find that the pre-trained initialization leads to higher generalization accuracy and consistent memorization rate of training data. These show that the pre-trained knowledge almost does not influence the LLMs' memorization of new fact knowledge and can improve LLMs' generalization on unseen facts. We leave further analyses of pre-trained influence to fact generalization as future work.
\section{Implementation Details}
\label{app:implementation_details}

In this section, we first introduce the general implementation details (model, training, dataset) and then introduce details of the specific settings for each individual experiment.
\subsection{Model Details}
We mainly use the model architecture and tokenizer
of Qwen-1.5~\citep{llm_qwen}. For more details of Qwen, we refer the reader to the original paper~\citep{llm_qwen}.
We mainly set the hyper-parameters of each LLM's size according to the following aspects:
1) The aspect ratio, which is the ratio of the hidden size to the number of layers, should be maintained at a moderate value. Following conventional design practices, we control the aspect ratio within the range of 128/3 (as adopted by Qwen-1.5-0.5B) to 128 (as adopted by Qwen-1.5-7B). 2) The intermediate size should be approximately 8/3 times of the hidden size and be divisible by 128. We provide the detailed hyper-parameters of model architecture in Table~\ref{tab:hyper_parameters_of_llms}.

\subsection{Training Details}
We configure the global batch size as 512 and employ the AdamW optimizer~\citep{optimizer_adam, optimizer_adamw}. 
In exploratory experiments, we find that LLMs with different sizes are highly sensitive to learning rates and thus we search for the best learning rates for each size's LLM and different datasets to achieve the optimal memorization rate, under small training epochs. Meanwhile, we adopt the cosine learning rate scheduler. We list the learning rates of each model size in Table~\ref{tab:hyper_parameters_of_llms}.
 It's observed that the optimal learning rates differed between the company information table and Wikidata, and the latter requires a higher learning rate. Most of these experiments are conducted using either 8 NVIDIA RTX 3090s or 4 NVIDIA A800s-80GB, utilizing BFloat16 mixed precision training. The training speed of models with different sizes can be referred to in Table~\ref{tab:training_speed}.

\subsection{Dataset Details}
In this paper, we conduct experiments on fact triples from a large real-world company information table and Wikidata\footnote{\href{https://www.wikidata.org/wiki/Wikidata:Introduction}{https://www.wikidata.org/wiki/Wikidata:Introduction}}.
The company information table is provided by a commercial data company and we obtain its permission to conduct this research. 
For Wikidata, we follow this public github repository\footnote{\href{https://github.com/neelguha/simple-wikidata-db}{https://github.com/neelguha/simple-wikidata-db}} to get all of its fact triples.
The facts of the company information table are in Chinese and Wikidata's facts are in English. 
For the key and entity in the company information table and Wikidata, we use their natural language name instead of the original key (uid) to closely mirror the facts in pre-training data.

\begin{table}[]
\small
\centering
\begin{tabular}{@{}lc@{}}
\toprule
\textbf{Model Identifier}   & \textbf{Training Speed} \\ 
\midrule
20M    &   800 \\
30M    &   700 \\
41M    &   650 \\
44M    &   600 \\
69M    &   500 \\
97M    &   400 \\
116M    &   300 \\
200M    &   225 \\
0.5B    &   100 \\
\bottomrule
\end{tabular}
\caption{Training speeds (triples per second per GPU) of models of different sizes, which are based on NVIDIA RTX 3090.}
\label{tab:training_speed}
\end{table}

\subsection{Details on Scaling Law of Fact Capacity and Model Size}
We conduct a series of experiments using various model sizes ranging from 30M to 0.5B (20M fails to reach 95\% MR at these epochs), while keeping the training epochs fixed. We use the number of non-embed parameters to measure model size, following \citet{scaling_law_for_lm}.
When randomly sampling facts, we first randomly sample keys and then use facts of these keys' all attributes to make fact type distributions consistent. In these experiments, we use $|D|*MR(D)$ to measure the fact capacity more accurately since the memorization rate may vary slightly for each $D$. 
The objective is to investigate the relationship between fact capacity and model size. Specifically, we utilize the results from models with fewer than 200M parameters to establish a scaling law formula. We then validate the extrapolation by employing models with 200M and 0.5B parameters on both the company information table and Wikidata. For Wikidata, we set the fixed training epochs to 100, while for the company information table, we use 50 and 200 epochs. In this way, we can observe the results across different datasets and epochs, which makes our results and the built scaling law more robust.
Besides, the templates we use for the company information table are listed in Table~\ref{tab:table_templates}. 
For Wikidata, since it contains tremendous types of fact (relations), it is costly to design an individual template for each type and thus we use a unified template: "For this entity, $\langle E \rangle$, the entity forming the relationship `$\langle R \rangle$' is:".

\subsection{Details on Scaling Law of Fact Capacity and Epochs}
To explore the relationship between fact capacity and memorization epochs, we conduct experiments using different quantities of fact triples from the company information table and train 44M/69M models to memorize these triples. To ensure convergence of loss and memorization rate, we set the maximum memorization epochs to be 1000 or even more. 
To save the computational cost, we manually stop the training once the model achieves a sufficiently high memorization rate (>95\%).
For each quantity of triples, we identify the first epoch in which the model attains a memorization rate higher than 95\%. We use this triple quantity as the memorization fact capacity at this epoch. After collecting these data points, we fit a negative exponential curve as shown in Figure \ref{fig:fact_capacity_and_epoch}. Furthermore, we observe that for quantities of triples exceeding the fact capacity saturation point, the model is unable to achieve a memorization rate higher than 95\% under 3000 training epochs, which almost reaches saturation.

\subsection{Details on Redundant Fact Memorization}
All of this section's experiments are conducted under 1000 epochs to ensure that the model's memorization reaches saturation.

For experiments on memorization of the same facts with different directions, we employ a 30M model and a 41M model to enhance model size diversity, and the triple quantity of each fact direction is 100K for the 30M model and 200K for the 41M model. The templates employed for forward and reverse versions of fact knowledge can be found in Table \ref{tab:direction_templates}.

For experiments on memorization of correlated facts with the same key, we employ a 20M model. 
To prevent the LLM from fully memorizing all facts (which makes the memorization rates of groups 100\% and hard to distinguish),
the number of triples for each attribute is set to be different, with 400K triples for attributes in group 2 (Title, Title-Code, Type, and Type-Code), and 100K triples for attributes in group 3 (Longitude, Start-Date, and Operator).

For experiments on memorization of derivable multi-hop facts, we employ a 30M model. The number of triples used for single-hop and two-hop knowledge is set to 200K. The templates employed for two-hop knowledge can be found in Table \ref{tab:two_hop_templates}.

For experiments on mixed training of fact memorization and abstract ability learning, we employ a 30M model to train on a combined dataset comprising 200K fact memorization triples and 200K samples from either the SNLI or Amazon-CLS train split. The templates utilized for SNLI and Amazon-CLS can be found in Table \ref{tab:abstract_ability_templates}.

\subsection{Details on Fact Memorization Preference}
Since this section focuses on memorization preference analysis, we select a combination of an attribute (Longitude or Operator) from the company information table and an attribute (Author) from Wikidata, to avoid the correlation between facts.
The model used for these experiments has a size of 30M, and the quantity of each type of fact knowledge (not triples) is set to 100K. To manipulate the frequency, we evenly up-sample one type of fact knowledge, thereby increasing its triple quantity.

For experiments investigating the difficulty preference of LLMs in memorization, we employ both a 30M model and a 41M model. The number of non-embed parameters in the 41M model is approximately twice that of the 30M model. In each experiment, we utilize 100K fact triples for each attribute (Longitude, Operator, and Credit-No).

For experiments investigating the memorization order preference of LLMs, we load a pre-trained checkpoint of a 30M model that has already trained to memorize 200K fact triples of one attribute. We then continue training this model to memorize an additional 200K fact triples of another attribute. This allow us to observe the influence of the memorization order on the model's performance.

\begin{figure}[]
    \centering

    \subfigure[Company $\to$ Credit-No]{

    \includegraphics[width=0.22\textwidth]{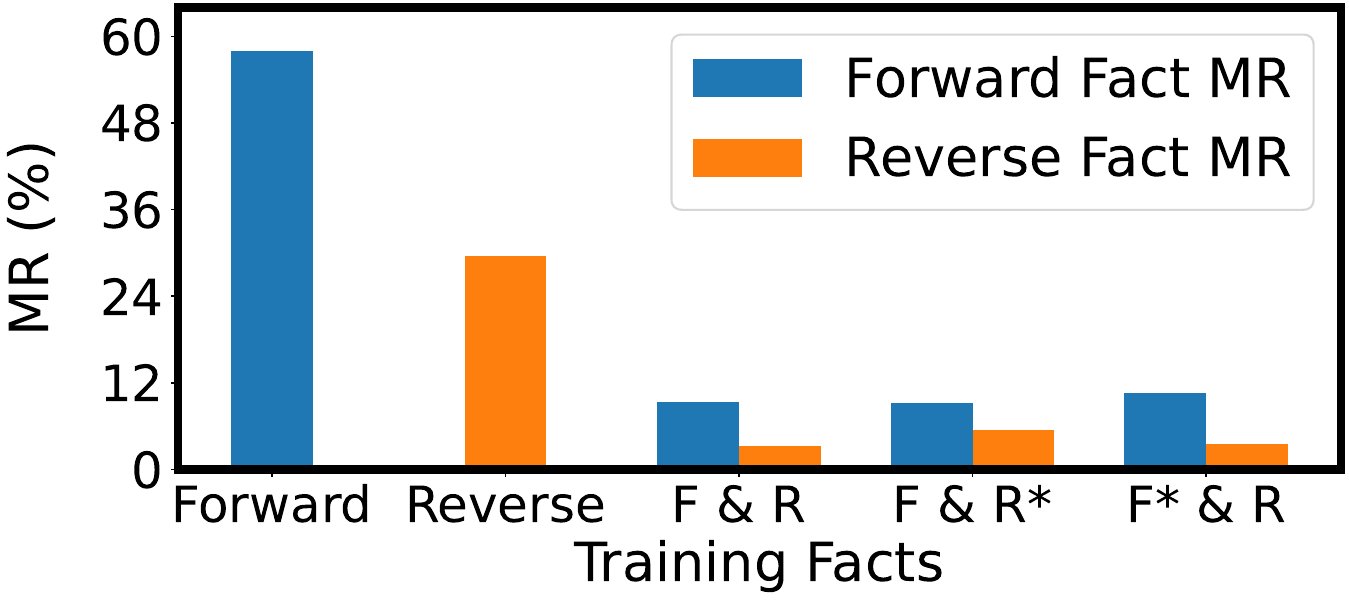}
    \includegraphics[width=0.215\textwidth]{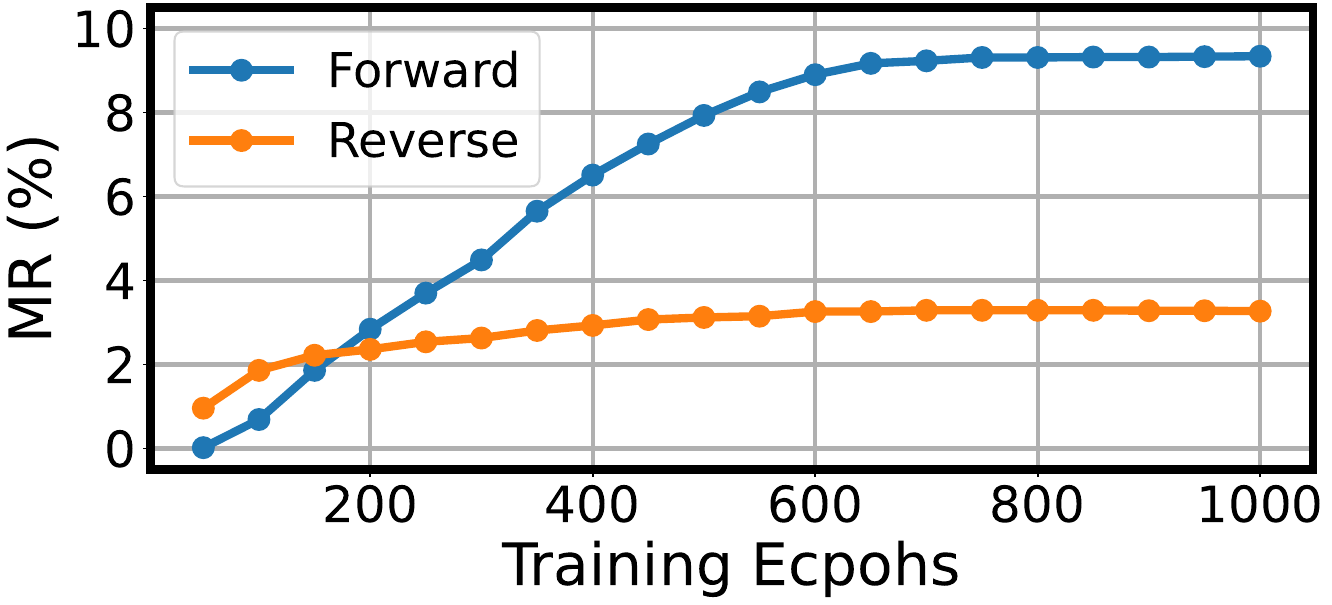}
    }

    \subfigure[Company $\to$ Operator]{

    \includegraphics[width=0.22\textwidth]{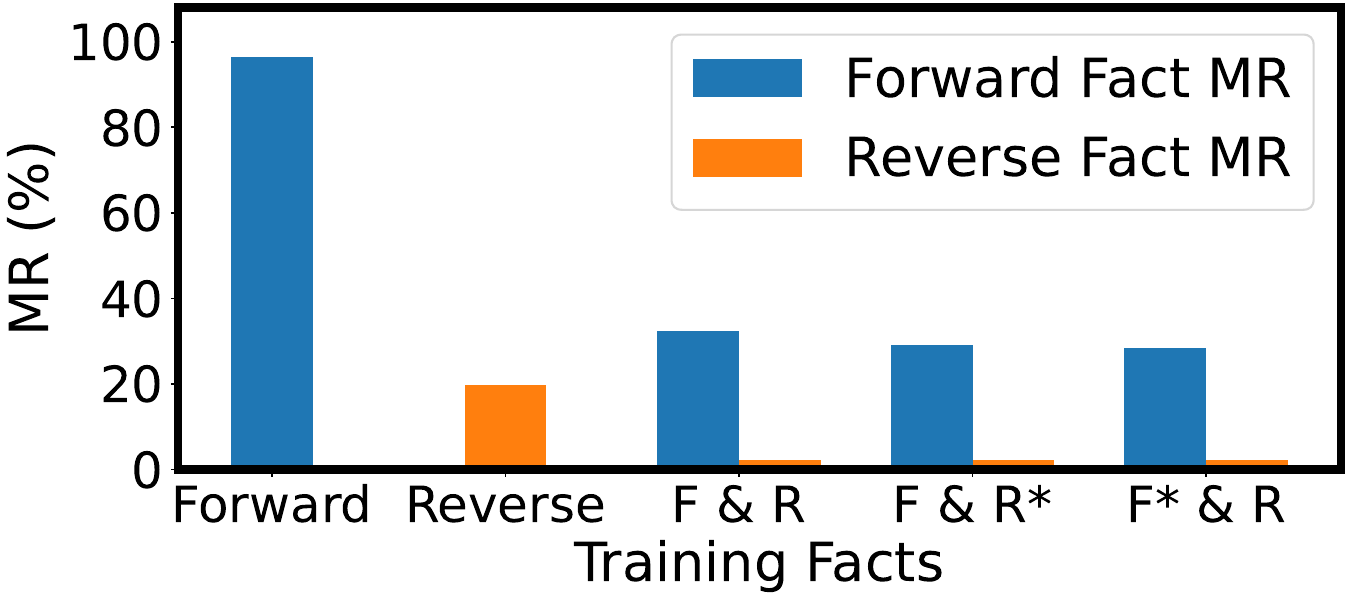}
    \includegraphics[width=0.215\textwidth]{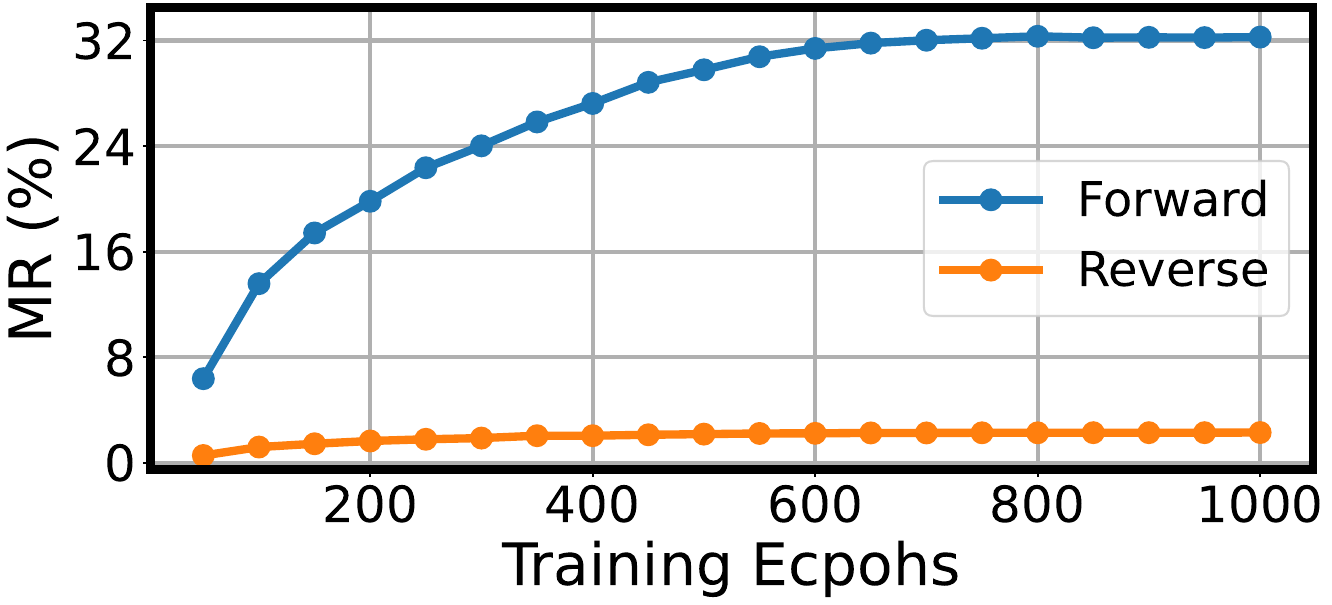}
    }

    \subfigure[Company $\to$ Register-No]{
    \includegraphics[width=0.22\textwidth]{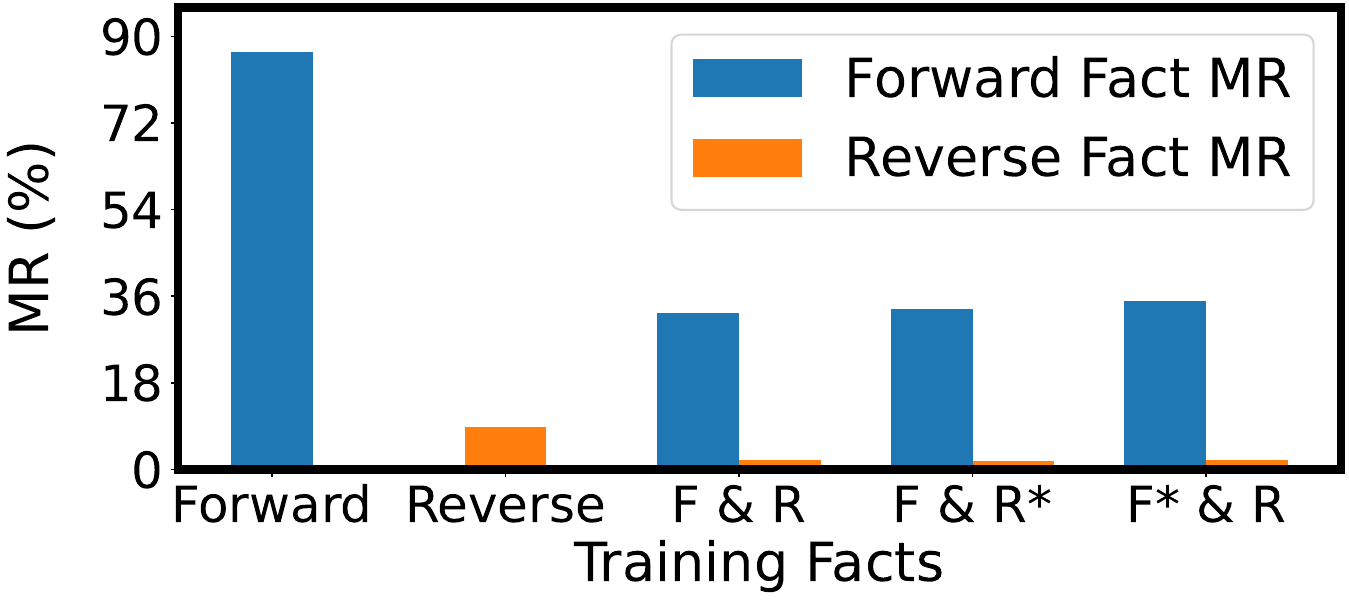}
    \includegraphics[width=0.215\textwidth]{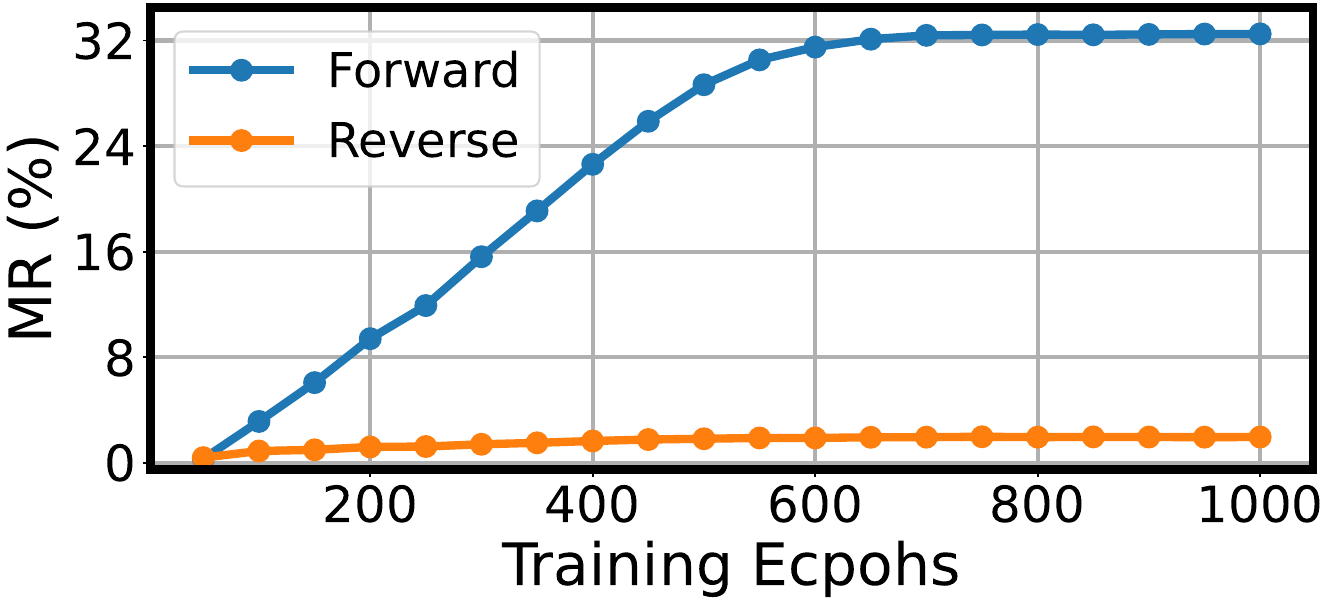}
    }
    \caption{LLMs' memorization of the same facts with different directions, on 30M model with 100K facts of each direction, where ``*'' means facts are from another group of keys. 
    The right side is the learning curves.
    }
    \label{fig:fact_compatibility_different_direction_30m}
    \vspace{-10pt}
\end{figure}

\section{Supplement Experiment for Memorizing Same Facts with Different Directions}
\label{app:fact_memorization_different_direction_30m}
We also analyze memorizing facts with different directions on the 30M model with 100K training facts and the results are shown in Figure~\ref{fig:fact_compatibility_different_direction_30m}, which exhibit the same trend as experiments of the 41M model.

\section{Supplement Experiment for Relation between Fact Memorization and Generalization}
\label{app:memorization_and_generalization_30m}
\begin{figure*}[t]
    \centering
    \includegraphics[width=0.7\textwidth]{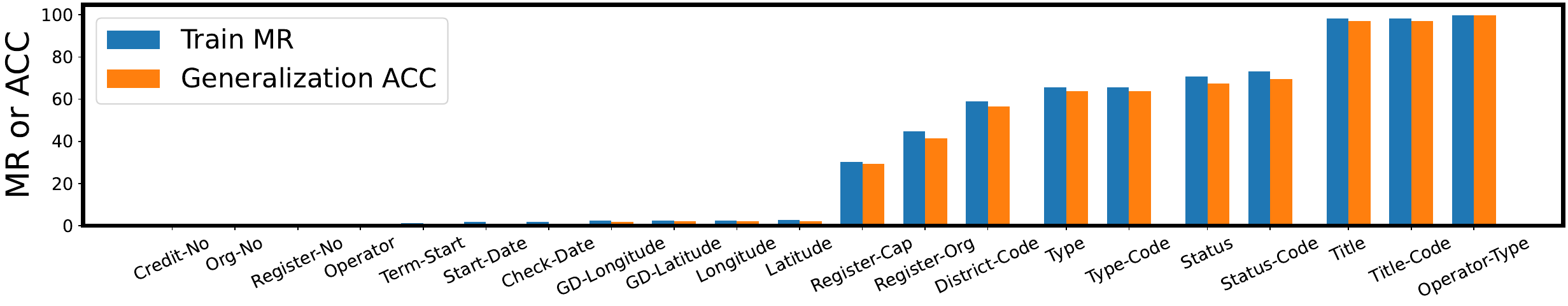}
    \vspace{-10pt}
    \caption{Memorization and generalization over facts of different types, on 30M LLM trained by 10M facts.}
    \vspace{-15pt}
    \label{fig:mr_test_acc_relation_30M}
\end{figure*}

We also analyze the relation between fact memorization and generalization on the 30M model with 10M training facts and the results are shown in Figure~\ref{fig:mr_test_acc_relation_30M}, which exhibit the same trend as experiments of the 44M model in FIgure~\ref{fig:mr_test_acc_relation_44M}.

\section{Generalization Case Studies}
\label{app:generalization_case_studies}
We show case studies of ``Company$\to$Longiude'', ``Company$\to$Capital'' and ``Company$\to$Type'' in Table~\ref{tab:case_study_company_longitude}, Table~\ref{tab:case_study_company_register_capital} and Table~\ref{tab:case_study_company_type}, respectively. 
The involved fact information is all publicly available from the official website.
\paragraph{Company$\to$Longitude} We find that LLMs may correctly predict the longitude of an unseen company according to the region name in the company name. Since the training facts contain the longitude of the company in ``Zhangjiajie'', the LLM can identify the association between the ``'Zhangjiajie'' and the longitude, and thus uses such association to predict another company in ``Zhangjiajie''. However, such association may lead to wrong prediction because the region name only can coarsely determine the rough range of the longitude. If a unseen company is very close to a training company with the same region name, the prediction will probably correct. Otherwise, the prediction may be wrong.
\paragraph{Company$\to$Capital}
Similar with ``Company $\to$Longitude'', the LLM can determine a rough range of unseen company's registered capital. The LLM can roughly estimate the company's register-capital according to company size indicated by the company name, e.g., ``Fruit shop''$\to$ (\textyen $10^4 \sim$ \textyen $10^5$) or ``Investment company''$\to$ (\textyen $10^7 \sim$ \textyen $10^9$). 
\paragraph{Company$\to$Type}
The LLM can easily judge a company's type based on the surface form of the company name.
\paragraph{Remark}
We find that LLMs' fact generalization depends on the correlation between input (key) and output (value)~\citep{short_cut_learning}. 
For a specific type of fact (attribute), the higher correlation between the key and value leads to higher generalization accuracy. 
For example, the LLM may correctly predict an unseen company's longitude if the company name contains a region name and the training dataset contains the longitude of companies with the same region name.
Or it can roughly estimate the company's register-capital according to company size indicated by the company name, e.g., ``Fruit shop''$\to$ (\textyen $10^4 \sim$ \textyen $10^5$) or ``Investment company''$\to$ (\textyen $10^7 \sim$ \textyen $10^9$). 
Meanwhile, different types of facts have different generalizability. For those facts with obvious patterns, the LLM can achieve reliable generalization.
These suggest the potential of adaptively leveraging LLM's fact generalization: 1. selectively leveraging generalization of those highly generalizable facts;
2. if the LLM does not exactly know the whole fact, it can response with a part of the fact, e.g., a rough range, to make its response more informative and thus helpful.

\begin{table*}[]
\small
\centering
\begin{tabular}{@{}lllc@{}}
\toprule
\textbf{Field}        & \textbf{Description}     & \textbf{Example}     & \textbf{Avg Tokens}        \\ \midrule
Company (Primary Key) & company name  &  Tiktok Co., Ltd.    & 13.2               \\
Credit-No        & social credit number  & 91110105MA...  & 13.5            \\
Operator              & legal representative     & Lidong Zhang  & 2.7      \\
Start-Date           & founding date             & 2003.11.2  & 10.0     \\
Title                 & representative title     & Executive Director   & 1.7  \\
Type                  & company type             & Co., Ltd.   & 7.6     \\
Longitude             & company longitude        & 116.497976   & 15.1     \\
Latitude              & company latitude         & 39.928384     & 14.5   \\
Register-No           & company registration number   & 4310271000119  & 14.5  \\
Organization-No        & company organization number   & 707414389   & 6.5  \\
Type-Code        & company type code        & 2190    & 4.0    \\
Title-Code            & representative title code   & 490A-Person in Charge  & 6.6 \\
Term-Start            & start date of the business term  & 2003.11.15 & 8.7 \\
Check-Date            & incorporation date       & 2006.12.04   & 9.7     \\
Register-Capital      & registered capital       & \textyen $10^5$   & 4.2     \\
Register-Org          & registration authority   & Shanghai AIC   & 6.1     \\
Operator-type         & the type of legal representative   & Individual & 1.0 \\
Status                & company status           & Open      & 7.5   \\
Status-Code           & company status code      & 0003  & 4.0    \\
GD-longitude          & company longitude on Amap
& 116.498  & 9.8 \\
GD-latitude           & company latitude on Amap & 39.928    & 8.9    \\
District-Code         & company district code
& 430182  & 6.0
\\ \bottomrule
\end{tabular}
\caption{All fields of the used large information table.}
\label{tab:table_all_fields}
\end{table*}

\begin{table*}[]
\small
\centering
\begin{tabular}{@{}lccc@{}}
\toprule
\textbf{Initialization} & \multicolumn{1}{l}{\textbf{Training Facts}} & \multicolumn{1}{l}{\textbf{MR}} & \multicolumn{1}{l}{\textbf{Generalization ACC}} \\ \midrule
Qwen-1.5-base-0.5B      & 4.3M                                        & 100                             & 32.64                                           \\
Random                  & 4.3M                                        & 100                             & 30.43                                           \\ \bottomrule
\end{tabular}
\caption{The comparison between pre-trained and random initialization, on 0.5B model.}
\label{tab:pre_train_comparison}
\end{table*}

\begin{table*}[]
\small
\centering
\begin{tabular}{@{}lccccccccc@{}}
\toprule
\textbf{Model Identifier}         & \textbf{20M} & \textbf{30M} & \textbf{41M} & \textbf{44M} & \textbf{69M} & \textbf{97M} & \textbf{116M} & \textbf{200M} & \textbf{0.5B} \\ \midrule
\textbf{All Parameters}           & 20.1M        & 30.5M        & 41.5M        & 44.0M        & 69.0M        & 97.1M        & 116.4M        & 201.6M        & 0.5B      \\
\textbf{Non-Embed Parameters} & 0.6M         & 1.3M         & 2.6M         & 5.1M         & 10.6M        & 19.3M        & 38.6M         & 85.0M         & 308M       \\
\textbf{Number of Layers}         & 3            & 3            & 3            & 6            & 6            & 6            & 12            & 24            & 24          \\
\textbf{Hidden Size}              & 128          & 192          & 256          & 256          & 384          & 512          & 512           & 768           & 1024       \\
\textbf{Intermediate Size}        & 384          & 512          & 768          & 768          & 1024         & 1408         & 1408          & 2048          & 2816       \\
\textbf{Attention Heads}          & 4            & 4            & 8            & 8            & 8            & 8            & 8             & 12            & 16        \\ 
\textbf{LR on Company Information Table}           
& 2.0e-3       & 1.0e-3        & 1.0e-3        & 7.5e-4        & 5.0e-4        & 5.0e-4        & 4.0e-4        & 2.5e-4        & 1.5e-4      \\
\textbf{LR on Wikidata}           
& 3.0e-3       & 2.0e-3        & 2.0e-3        & 1.5e-3        & 1.0e-3        & 7.5e-4        & 7.5e-4        & 5.0e-4        & 3.0e-4      \\
\bottomrule
\end{tabular}
\caption{Hyper-parameters of LLMs with different sizes.}
\label{tab:hyper_parameters_of_llms}
\end{table*}

\begin{table*}[]
\centering
\small
\begin{tabular}{@{}ll@{}}
\toprule
\textbf{Attribute}  & \textbf{Template} \\ 
\midrule
\multirow{2}{*}{Credit-No}  & \begin{CJK}{UTF8}{gbsn}在企业基本信息表中，公司：“$\langle C \rangle$”的“社会信用号”为：\end{CJK} \\
& (In the company information table, the "Credit-No" of the company "$\langle C \rangle$" is:)\\ \midrule
\multirow{2}{*}{Operator}  & \begin{CJK}{UTF8}{gbsn}在企业基本信息表中，公司：“$\langle C \rangle$”的“法定代表人”为：\end{CJK} \\
& (In the company information table, the "Operator" of the company "$\langle C \rangle$" is:)\\ 
\midrule
\multirow{2}{*}{Start-Date}  & \begin{CJK}{UTF8}{gbsn}在企业基本信息表中，公司：“$\langle C \rangle$”的“成立日期”为：\end{CJK} \\
& (In the company information table, the "Star-Date" of the company "$\langle C \rangle$" is:)\\ 
\midrule
\multirow{2}{*}{Title}  & \begin{CJK}{UTF8}{gbsn}在企业基本信息表中，公司：“$\langle C \rangle$”的“公司代表人职务”为：\end{CJK} \\
& (In the company information table, the "Title" of the company "$\langle C \rangle$" is:)\\ 
\midrule
\multirow{2}{*}{Title}  & \begin{CJK}{UTF8}{gbsn}在企业基本信息表中，公司：“$\langle C \rangle$”的“企业类型”为：\end{CJK} \\
& (In the company information table, the "Type" of the company "$\langle C \rangle$" is:)\\ 
\midrule
\multirow{2}{*}{Longitude}  & \begin{CJK}{UTF8}{gbsn}在企业基本信息表中，公司：“$\langle C \rangle$”的“经度”为：\end{CJK} \\
& (In the company information table, the "Longitude" of the company "$\langle C \rangle$" is:)\\ 
\midrule
\multirow{2}{*}{Latitude}  & \begin{CJK}{UTF8}{gbsn}在企业基本信息表中，公司：“$\langle C \rangle$”的“纬度”为：\end{CJK} \\
& (In the company information table, the "Latitude" of the company "$\langle C \rangle$" is:)\\ 
\midrule
\multirow{2}{*}{Register-No}  & \begin{CJK}{UTF8}{gbsn}在企业基本信息表中，公司：“$\langle C \rangle$”的“注册号”为：\end{CJK} \\
& (In the company information table, the "Register-No" of the company "$\langle C \rangle$" is:)\\ 
\midrule
\multirow{2}{*}{Organization-No}  & \begin{CJK}{UTF8}{gbsn}在企业基本信息表中，公司：“$\langle C \rangle$”的“组织机构号”为：\end{CJK} \\
& (In the company information table, the "Organization-No" of the company "$\langle C \rangle$" is:)\\ 
\midrule
\multirow{2}{*}{Type-Code}  & \begin{CJK}{UTF8}{gbsn}在企业基本信息表中，公司：“$\langle C \rangle$”的“企业类型代码”为：\end{CJK} \\
& (In the company information table, the "Type-Code" of the company "$\langle C \rangle$" is:)\\ 
\midrule
\multirow{2}{*}{Title-Code}  & \begin{CJK}{UTF8}{gbsn}在企业基本信息表中，公司：“$\langle C \rangle$”的“代表人类型代码”为：\end{CJK} \\
& (In the company information table, the "Title-Code" of the company "$\langle C \rangle$" is:)\\ 
\midrule
\multirow{2}{*}{Term-Start}  & \begin{CJK}{UTF8}{gbsn}在企业基本信息表中，公司：“$\langle C \rangle$”的“经营期限起始日期”为：\end{CJK} \\
& (In the company information table, the "Term-Start" of the company "$\langle C \rangle$" is:)\\ 
\midrule
\multirow{2}{*}{Check-Date}  & \begin{CJK}{UTF8}{gbsn}在企业基本信息表中，公司：“$\langle C \rangle$”的“核准日期”为：\end{CJK} \\
& (In the company information table, the "Check-Date" of the company "$\langle C \rangle$" is:)\\ 
\midrule
\multirow{2}{*}{Register-Capital}  & \begin{CJK}{UTF8}{gbsn}在企业基本信息表中，公司：“$\langle C \rangle$”的“注册资本”为：\end{CJK} \\
& (In the company information table, the "Register-Capital" of the company "$\langle C \rangle$" is:)\\ 
\midrule
\multirow{2}{*}{Register-Org}  & \begin{CJK}{UTF8}{gbsn}在企业基本信息表中，公司：“$\langle C \rangle$”的“登记机关”为：\end{CJK} \\
& (In the company information table, the "Register-Org" of the company "$\langle C \rangle$" is:)\\ 
\midrule
\multirow{2}{*}{Operator-type}  & \begin{CJK}{UTF8}{gbsn}在企业基本信息表中，公司：“$\langle C \rangle$”的“代表人类型代码”为：\end{CJK} \\
& (In the company information table, the "Operator-type" of the company "$\langle C \rangle$" is:)\\ 
\midrule
\multirow{2}{*}{Status}  & \begin{CJK}{UTF8}{gbsn}在企业基本信息表中，公司：“$\langle C \rangle$”的“状态”为：\end{CJK} \\
& (In the company information table, the "Status" of the company "$\langle C \rangle$" is:)\\ 
\midrule
\multirow{2}{*}{Status-Code}  & \begin{CJK}{UTF8}{gbsn}在企业基本信息表中，公司：“$\langle C \rangle$”的“企业状态码”为：\end{CJK} \\
& (In the company information table, the "Status-Code" of the company "$\langle C \rangle$" is:)\\ 
\midrule
\multirow{2}{*}{GD-Longitude}  & \begin{CJK}{UTF8}{gbsn}在企业基本信息表中，公司：“$\langle C \rangle$”在高德地图上的“经度”为：\end{CJK} \\
& (In the company information table, the "GD-Longitude" of the company "$\langle C \rangle$" is:)\\ 
\midrule
\multirow{2}{*}{GD-Latitude}  & \begin{CJK}{UTF8}{gbsn}在企业基本信息表中，公司：“$\langle C \rangle$”在高德地图上的“纬度”为：\end{CJK} \\
& (In the company information table, the "GD-Latitude" of the company "$\langle C \rangle$" is:)\\ 
\midrule
\multirow{2}{*}{District-Code}  & \begin{CJK}{UTF8}{gbsn}在企业基本信息表中，公司：“$\langle C \rangle$”的“区域码”为：\end{CJK} \\
& (In the company information table, the "District-Code" of the company "$\langle C \rangle$" is:)\\ 
\bottomrule
\end{tabular}
\caption{Templates of each attribute for the company information table memorization.}
\label{tab:table_templates}
\end{table*}

\begin{table*}[]
\centering
\small
\begin{tabular}{@{}lll@{}}
\toprule
\textbf{Attribute}  & \textbf{Direction}    & \textbf{Template} \\ 
\midrule
\multirow{4}{*}{Credit-No} & \multirow{2}{*}{Forward}  & \begin{CJK}{UTF8}{gbsn}在企业基本信息表中，公司：“$\langle C \rangle$”的“社会信用号”为：\end{CJK} \\
& & (In the company information table, the "Credit-No" of the company "$\langle C \rangle$" is:) \\
& \multirow{2}{*}{Reverse}  & \begin{CJK}{UTF8}{gbsn}在企业基本信息表中，“社会信用号”是$\langle CNo \rangle$”的公司为：\end{CJK} \\
& & (In the company information table, the company with the "Credit-No" as $\langle CNo \rangle$ is:) \\
\midrule
\multirow{4}{*}{Operator} & \multirow{2}{*}{Forward}  & \begin{CJK}{UTF8}{gbsn}在企业基本信息表中，公司：“$\langle C \rangle$”的“法定代表人”为：\end{CJK} \\
& & (In the company information table, the "Operator" of the company "$\langle C \rangle$" is:) \\
& \multirow{2}{*}{Reverse}  & \begin{CJK}{UTF8}{gbsn}在企业基本信息表中，“法定代表人”是$\langle Op \rangle$”的公司为：\end{CJK} \\
& & (In the company information table, the company with the "Operator" as $\langle Op \rangle$ is:) \\
\midrule
\multirow{4}{*}{Register-No} & \multirow{2}{*}{Forward}  & \begin{CJK}{UTF8}{gbsn}在企业基本信息表中，公司：“$\langle C \rangle$”的“注册号”为：\end{CJK} \\
& & (In the company information table, the "Register-No" of the company "$\langle C \rangle$" is:) \\
& \multirow{2}{*}{Reverse}  & \begin{CJK}{UTF8}{gbsn}在企业基本信息表中，“注册号”是$\langle RNo \rangle$”的公司为：\end{CJK} \\
& & (In the company information table, the company with the "Register-No" as $\langle RNo \rangle$ is:) \\
\bottomrule
\end{tabular}
\caption{Templates of forward and reverse version of fact knowledge memorization.}
\label{tab:direction_templates}
\end{table*}

\begin{table*}[]
\centering
\small
\begin{tabular}{@{}ll@{}}
\toprule
\textbf{Attribute}  & \textbf{Template} \\ 
\midrule
\multirow{2}{*}{Longitude}  & \begin{CJK}{UTF8}{gbsn}在企业基本信息表中，“$\langle C_A \rangle$”与“$\langle C_B \rangle$”在“经度”上的差值为：\end{CJK} \\
& (In the company information table, the difference in "Longitude" between "$\langle C_A \rangle$" and "$\langle C_B \rangle$" is:)\\ 
\midrule
\multirow{2}{*}{Start-Date}  & \begin{CJK}{UTF8}{gbsn}在企业基本信息表中，“$\langle C_A \rangle$”与“$\langle C_B \rangle$”的“成立日期”相差：\end{CJK} \\
& (In the company information table, the difference in "Start-Date" between "$\langle C_A \rangle$" and "$\langle C_B \rangle$" is: )\\ 
\bottomrule
\end{tabular}
\caption{Templates of derivable two-hop fact knowledge memorization.}
\label{tab:two_hop_templates}
\end{table*}

\begin{table*}[]
\centering
\small
\begin{tabular}{@{}ll@{}}
\toprule
\textbf{Dataset}  & \textbf{Template} \\ 
\midrule
SNLI    &   Premise: $\langle Premise \rangle$ \textbackslash n Hypothesis: $\langle Hypothesis \rangle$ \textbackslash n The relation between the premise and the hypothesis is: \\
\midrule
Amazon-CLS    &  What is the rating of the following amazon review: \textbackslash n review title: $\langle Title \rangle$ \textbackslash n review content: $\langle Title \rangle$ \textbackslash n rating: \\
\bottomrule
\end{tabular}
\caption{Templates of abstract ability learning.}
\label{tab:abstract_ability_templates}
\end{table*}

\begin{table*}[]
\small
\centering
\begin{tabular}{@{}lccc@{}}
\toprule
\textbf{Train/Test} & \textbf{Company}                                      & \textbf{Prediction} & \textbf{Gold}      \\ \midrule
Train               & Zhangjiajie Natural Agriculture Development Co., Ltd. & 110.4939900034971   & 110.4939900034971  \\
Test                & Zhangjiajie Jiahao Construction Engineering Co., Ltd. & 110.4939900034971   & 110.4939900034971  \\
Test                & Zhangjiajie Changtu Construction Co., Ltd.            & 110.4939900034971   & 110.490945197      \\
Test                & Zhangjiajie Yiming Life Supermarket Co., Ltd.         & 110.4939900034971   & 110.48127269239656 \\ \bottomrule
\end{tabular}
\caption{Case study on fact generalization on Company$\to$Longitude.}
\label{tab:case_study_company_longitude}
\vspace{20pt}
\begin{tabular}{@{}lccc@{}}
\toprule
\textbf{Train/Test} & \textbf{Company}                                                     & \textbf{Prediction}       & \textbf{Gold}             \\ \midrule
Train               & Jishou City Fruit Shop                                               & \textyen 10\textasciicircum{}4 & \textyen 10\textasciicircum{}4 \\
Test                & Yongzhou City Handsome Sister Fruit Shop                             & \textyen 10\textasciicircum{}4 & \textyen 10\textasciicircum{}4 \\
Test                & Yueyang City Chenghong Fruit Shop                                    & \textyen 10\textasciicircum{}4 & \textyen 10\textasciicircum{}5  \\
Train               & Beijing Guojintan Asset Management Co., Ltd.                         & \textyen 10\textasciicircum{}8 & \textyen 10\textasciicircum{}8  \\
Test                & Hunan Diamond Financing Guarantee Co., Ltd.                          & \textyen 10\textasciicircum{}8  & \textyen 10\textasciicircum{}8  \\
Test                & Xiangtan Urban Development Investment and Management Group Co., Ltd. & \textyen 10\textasciicircum{}8 & \textyen 10\textasciicircum{}9 \\ \bottomrule
\end{tabular}
\caption{Case study on fact generalization on Company$\to$Register-Capital.}
\label{tab:case_study_company_register_capital}
\vspace{20pt}
\begin{tabular}{@{}lccc@{}}
\toprule
\textbf{Train/Test} & \textbf{Company}                                     & \textbf{Prediction} & \textbf{Gold}       \\ \midrule
Test                & Changsha Yuyun Real Estate Brokerage Co., Ltd.       & Co., Ltd.           & Co., Ltd.           \\
Test                & Beijing Shenchen Information Technology Co., Ltd.    & Co., Ltd.           & Co., Ltd.           \\
Test                & Hunan Chuangneng Investment Co., Ltd.                & Co., Ltd.           & Co., Ltd.           \\
Test                & Hunan Zhongtie Travel Agency Co., Ltd. Lusong Branch & LLC Branch          & LLC Branch          \\
Test                & Yueyang Jiulong Supermarket Co., Ltd. Nanhu Branch   & LLC Branch          & LLC Branch          \\
Test                & Changsha Tongshan Department Store                   & Sole Proprietorship & Sole Proprietorship \\
Test                & Xiangcheng Hotel, Taoyuan County                     & Sole Proprietorship & Sole Proprietorship \\ \bottomrule
\end{tabular}
\caption{Case study on fact generalization on Company$\to$Type.}
\label{tab:case_study_company_type}
\end{table*}

\end{document}